\documentclass[twoside,11pt]{article}

\usepackage{jmlr2e}
\usepackage{amsmath}

\ShortHeadings{A Survey on Multi-view Learning}{}
\firstpageno{1}

\begin{document}

\title{A Survey on Multi-view Learning}

\author{\name Chang Xu \email changxu1989@gmail.com \\
        \addr  Centre for Quantum Computation and Intelligent Systems\\
       Faculty of Engineering and Information Technology\\
       University of Technology, Sydney\\
       Sydney, NSW 2007, Australia
       \AND
       \name Dacheng Tao \email dacheng.tao@uts.edu.au \\
       \addr  Centre for Quantum Computation and Intelligent Systems\\
       Faculty of Engineering and Information Technology\\
       University of Technology, Sydney\\
       Sydney, NSW 2007, Australia
       \AND
       \name Chao Xu \email xuchao@cis.pku.edu.cn\\
       \addr  Key Laboratory of Machine Perception (Ministry of Education)\\
       School of Electronics Engineering and Computer Science\\
       Peking University\\
        Beijing 100871, China
}


\maketitle

\begin{abstract}
In recent years, a great many methods of learning from multi-view data by considering the diversity of different views have been proposed. These views may be obtained from multiple sources or different feature subsets. For example, a person can be identified by face, fingerprint, signature or iris with information obtained from multiple sources, while an image can be represented by its color or texture features, which can be seen as different feature subsets of the image. In trying to organize and highlight similarities and differences between the variety of multi-view learning approaches, we review a number of representative multi-view learning algorithms in different areas and classify them into three groups: 1) co-training, 2) multiple kernel learning, and 3) subspace learning. Notably, co-training style algorithms train alternately to maximize the mutual agreement on two distinct views of the data; multiple kernel learning algorithms exploit kernels that naturally correspond to different views and combine kernels either linearly or non-linearly to improve learning performance; and subspace learning algorithms aim to obtain a latent subspace shared by multiple views by assuming that the input views are generated from this latent subspace. Though there is significant variance in the approaches to integrating multiple views to improve learning performance, they mainly exploit either the consensus principle or the complementary principle to ensure the success of multi-view learning. Since accessing multiple views is the fundament of multi-view learning, with the exception of study on learning a model from multiple views, it is also valuable to study how to construct multiple views and how to evaluate these views. Overall, by exploring the consistency and complementary properties of different views, multi-view learning is rendered more effective,  more promising, and has better generalization ability than single-view learning.
\end{abstract}


\section{Introduction}

\begin{figure*}[th]
\begin{center}
\includegraphics[width=\textwidth]{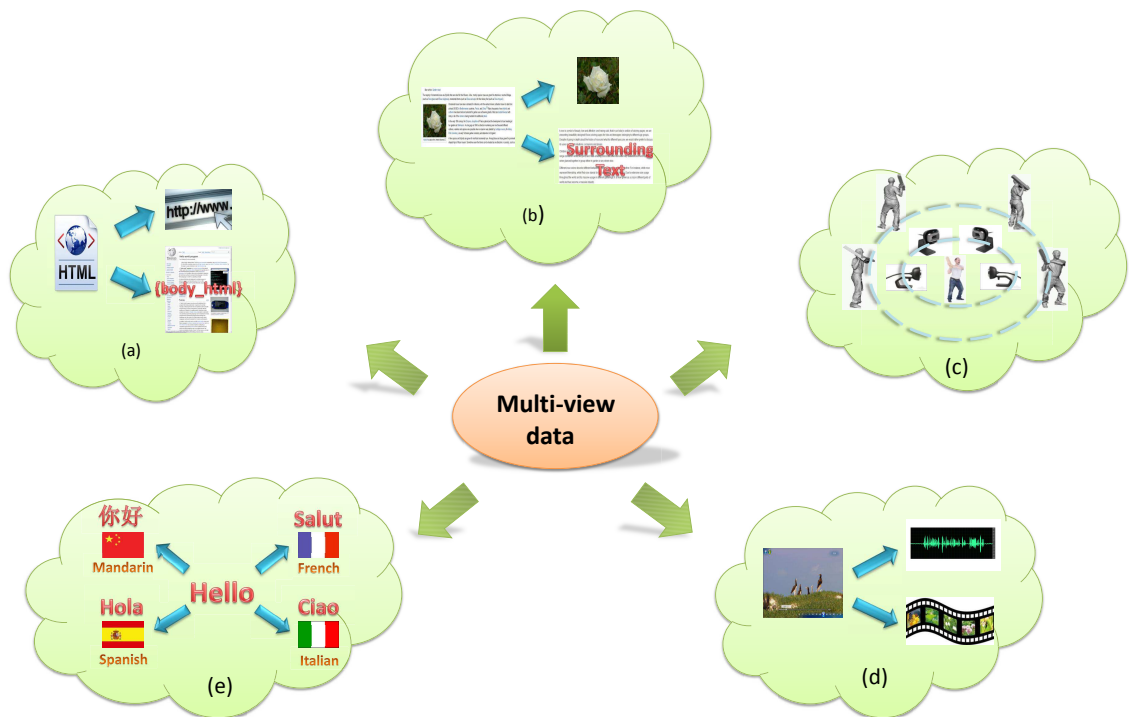}
\caption{Multi-view data: a) a web document can be represented by its url and words on the page, b) a web image can be depicted by its surrounding text separate to the visual information, c) images of a 3D object taken from different viewpoints, d) video clips are combinations of audio signals and visual frames, e) multilingual documents have one view in each language.}\vspace{-5mm}
\end{center}
\label{fig:1}
\end{figure*}

In most scientific data analytics problems in video surveillance, social computing, and environmental sciences, data are collected from diverse domains or obtained from various feature extractors and exhibit heterogeneous properties, because variables of each data example can be naturally partitioned into groups. Each variable group is referred to as a particular view, and the multiple views for a particular problem can take different forms, e.g. a) colour descriptor, local binary patterns, local shape descriptor, slow features and spatial temporal context captured by multiple cameras for person re-identification and global activity understanding in sparse camera network, and b) words in documents, information describing documents (e.g. title, author and journal) and the co-citation network graph for scientific document management (see Figure \ref{fig:1}).

Conventional machine learning algorithms, such as support vector machines, discriminant analysis, kernel machines, and spectral clustering, concatenate all multiple views into one single view to adapt to the learning setting. However, this concatenation causes over-fitting in the case of a small size training sample and is not physically meaningful because each view has a specific statistical property. In contrast to single view learning, multi-view learning as a new paradigm introduces one function to model a particular view and jointly optimizes all the functions to exploit the redundant views of the same input data and improve the learning performance. Therefore, multi-view learning has been receiving increased attention and existing algorithms can be classified into three groups: 1) co-training, 2) multiple kernel learning, and 3) subspace learning.

Co-training~\citep{Co-training} is one of the earliest schemes for multi-view learning. It trains alternately to maximize the mutual agreement on two distinct views of the unlabeled data. Many variants have since been developed. \cite{Co-em} generalized expectation-maximization (EM) by assigning changeable probabilistic labels to unlabeled data. \cite{muslea2002active+, muslea2003active, muslea2006active} combined active learning with co-training and proposed robust semi-supervised learning algorithms. \cite{yu:bayesian, yu2011bayesian} developed a Bayesian undirected graphical model for co-training and a novel co-training kernel for Gaussian process classifiers. \cite{Co-traning-label} treated co-training as the combinative label propagation over two views and unified the graph- and disagreement-based semi-supervised learning into one framework. \cite{Co-regularization} constructed a data-dependent ``co-regularization'' norm. The resultant reproducing kernel associated with a single RKHS simplified the theoretical analysis and extended the algorithmic scope of co-regularization. \cite{multi_view_clustering} and \cite{kumar2010co,Co-regularization-clustering} advanced co-training for data clustering and designed effective algorithms for multi-view data. The success of co-training algorithms mainly relies on three assumptions: (a) sufficiency - each view is sufficient for classification on its own, (b) compatibility- the target function of both views predict the same labels for co-occurring features with a high probability, and (c) conditional independence- views are conditionally independent given the label. The conditional independence assumption is critical, but it is usually too strong to satisfy in practice and thus several weaker alternatives~\citep{bootstrapping, Co-training-expansion, Co-training-difference} have been considered.

Multiple kernel learning (MKL) was originally developed to control the search space capacity of possible kernel matrices to achieve good generalization but has been widely applied to problems involving multi-view data. This is because kernels in MKL naturally correspond to different views and combining kernels either linearly or non-linearly improves learning performance. \cite{lanckriet2002learning, lanckriet2004learning} formulated MKL as a semi-definite programming problem. \cite{bach2004multiple}  treated MKL as a second order cone program problem and developed an SMO algorithm to efficiently obtain the optimal solution. \cite{sonnenburg2006general, sonnenburg2006large} developed an efficient semi-infinite linear program and made MKL applicable to large scale problems. \cite{rakotomamonjy2007more, rakotomamonjy2008simplemkl} proposed simple MKL by exploring an adaptive 2-norm regularization formulation. \cite{szafranski2008composite, szafranski2010composite, xu2010simple} and \cite{subrahmanya2010sparse} constructed the connection between MKL and group-LASSO to model group structure. Many generalization bounds have been obtained to theoretically guarantee the performance of MKL. \cite{lanckriet2004learning} showed that given $k$ base kernels, the estimation error is bounded by $O(\sqrt{\frac{k/\gamma^{2}}{n}})$, where $\gamma$ is the margin of the learned classifier. \cite{ying2009generalization} used the metric entropy integrals and pseudo-dimension of a set of candidate kernels to estimate the empirical Rademacher chaos complexity. The generalization bounds have a logarithmic dependency on $k$ for the family of convex combinations of $k$ base kernels with the $l_{1}$ constraint. Assuming different views to be uncorrelated, \cite{kloft2011local} derived a tighter upper bound by the local Rademacher complexities for the $l_{p}$-norm MKL. The cited survey \citep{gonen2011multiple} is believed to contain all the related references omitted from the proposal.

Subspace learning-based approaches aim to obtain a latent subspace shared by multiple views by assuming that the input views are generated from this latent subspace. The dimensionality of the latent subspace is lower than that of any input view, so subspace learning is effective in reducing the ``curse of dimensionality''. Given this subspace, it is straightforward to conduct the subsequent tasks, such as classification and clustering. Canonical correlation analysis (CCA) \citep{hotelling1936relations} and kernel canonical correlation analysis (KCCA) \citep{akaho2006kernel} explore basis vectors for two sets of variables by mutually maximizing the correlations between the projections onto these basis vectors, so it is straightforward to apply them to two-view data to select the shared latent subspace. They have been further developed to conduct multi-view clustering \citep{muliview_clustering_cca} and regression \citep{co-regression-cca}. \cite{fda} generalized Fisher's discriminant analysis to explore the latent subspace spanned by multi-view data. In contrast to CCA, this generalization considers the class label information. \cite{quadrianto2011learning} and \cite{zhai2012multiview} studied multi-view metric learning by constructing embedding projections from multi-view data to a shared subspace, where the Euclidean distance is meaningful across different views. The latent subspace is valuable for inferring another view from the observation view. \cite{sgplvm} exploited Gaussian process, \cite{skie} maximized the mutual information, and \cite{multi_view_mn} used Markov network to construct the connections between the two views through latent subspaces. \cite{fols} and \cite{fls_sparsity} proposed to find a latent subspace in which the information is correctly factorized into shared and private parts across different views. Consistency and finite sample analysis \citep{fukumizu2007statistical, hardoon2009convergence, cai2011convergence} have been studied for KCCA.

In reviewing the literature on multi-view learning, we find it is tightly connected with other topics in machine learning, such as active learning, ensemble learning and domain adaptation. Active learning \citep{settles.tr09, seung1992query}, sometimes called query learning, aims to minimize the amount of labeled data required for learning a concept of interest. \cite{muslea2000selective} introduced co-testing, which is a novel approach to conducting active learning with multiple views. They combined co-testing with co-EM and derived a novel method co-EMT \citep{muslea2002active+}, which uses co-EM to generate accurate classifiers and chooses the most informative unlabeled examples for co-Testing to label. Furthermore, considering strong and weak views, \cite{muslea2003active, muslea2006active} advanced co-Testing by assuming that the concentrated examples whose labels from strong classifiers are different and inconsistent with the prediction of the weak classifier provide more information for labeling. The idea of ensemble learning \citep{dietterichl2002ensemble, lappalainen2000ensemble} is to employ multiple learners and combine their predictions. The bagging algorithm \citep{breiman1996bagging} uses different training datasets to construct each member of the ensemble and predicts through uniform averaging or voting over class labels. In contrast to co-training, which ensures the diversities of the learned models by training on distinct views, bagging requires different training datasets for generating models with different judgments. AdaBoost \citep{freund1996experiments} is another well-known ensemble learning algorithm, in which the principal idea is to train a new model to compensate for the errors made by earlier models. In each round, the misclassified examples are identified and their emphasis will be increased in a new training set for the next training process. Both co-training and AdaBoost rely on a growing ensemble of classifiers trained on resamples of the data; however, AdaBoost tries to find the error-labeled examples, whereas co-training attempts to exploit the agreement of the learners. Co-training is confidence-driven whereas AdaBoost is error-driven. Domain adaptation refers to the problem of adapting a prediction model trained on data from a source domain to a different target domain, where the data distributions in the two domains are different. Many domain adaptation techniques \citep{wei2010cross, wan2011bi} have been proposed to solve the cross-language text classification problem, where the source domain includes the documents translated from the source language and the target domain includes the original documents in the target language. Moreover, these documents in different languages can be seen as different views of the original document; thus, methods like co-training \citep{wan2009co}, multi-view majority voting \citep{amini2010learning} and multi-view co-classification \citep{amini2010co} have been designed and successfully applied for this problem.

In this survey paper, we provide a comprehensive overview of multi-view learning. The rest of this paper is organized as follows: we first illustrate the principles underlying multi-view learning algorithms in Section \ref{sec:2}. In Section \ref{sec:3}, different approaches to the construction of multiple views and methods to evaluate these views are introduced. We present different ways to combine multiple views in Section \ref{sec:4} and illustrate different kinds of multi-view learning algorithm in detail in Sections \ref{sec:5}, \ref{sec:6} and \ref{sec:7}. The applications of multi-view learning are introduced in Section \ref{sec:8} and experimental results reflecting the performance of multi-view learning are shown in Section \ref{sec:9}. Finally, we conclude the paper in Section \ref{sec:10}.

\section{Principles for Multi-view Learning}
\label{sec:2}

The demand for redundant views of the same input data is a major difference between multi-view and single-view learning algorithms. Thanks to these multiple views, the learning task can be conducted with abundant information. However if the learning method is unable to cope appropriately with multiple views, these views may even degrade the performance of multi-view learning. Through fully considering the relationships between multiple views, several successful multi-view learning techniques have been proposed. We analyze these various algorithms and observe that there are two significant principles ensuring their success: \emph{consensus} and \emph{complementary} principles.

\subsection{Consensus Principle}

Consensus principle aims to maximize the agreement on multiple distinct views. Suppose the available data set $X$ has two views $X^{1}$ and $X^{2}$. An example $(x_{i},y_{i})$ is therefore viewed as $(x_{i}^{1},x_{i}^{2},y_{i})$, where $y_{i}$ is the label associated with the example. \cite{dasgupta2002pac} demonstrated the connection between the consensus of two hypotheses on two views respectively and their error rates. Under some mild assumptions they gave the inequality
\begin{equation}\label{}\nonumber
    P(f^{1}\neq f^{2})\geq \max\{P_{err}(f^{1}),P_{err}(f^{2})\}.
\end{equation}
From the inequality, we conclude that the probability of a disagreement of two independent hypotheses upper bounds the error rate of either hypothesis. Thus by minimizing the disagreement rate of the two hypotheses, the error rate of each hypothesis will be minimized.

In recent years, a great number of methods appear to have utilized this consensus principle in one way or another, even though in many cases the contributors are not aware of the relationship between their methods and this common underlying principle. For example, the co-training algorithm trains alternately to maximize the mutual agreement on two distinct views of the unlabeled data. By minimizing the error on labeled examples and maximizing the agreement on unlabeled examples, the co-training algorithm finally achieves one accurate classifier on each view. In the co-regularization algorithm, the consensus principle can be formulated by regularization terms as
\begin{equation}\label{Co-regularization}
    \min \quad \sum_{i\in U}[f^{1}(x_{i})-f^{2}(x_{i})]^{2}+\sum_{i\in L}V(y_{i},f(x_{i})),
\end{equation}
where the first term enforces the agreement on two distinct views on unlabeled examples, and the second term evaluates the empirical loss on the labeled examples with respect to a loss function $V(\cdot,\cdot)$. By additionally considering the complexity of the hypotheses, we will achieve the complete objective function, and solving it will result in learning two optimal hypotheses. Observing that applying the kernel canonical correlation analysis (KCCA) to the two feature spaces can improve the performance of the classifier, \cite{farquhar2005two} proposed a supervised learning algorithm called SVM-2K, which combines the idea of KCCA with SVM. An SVM can be thought of as projecting the feature to a 1-dimensional space followed by thresholding, after which SVM-2K forces the constraint of consensus of two views on this 1-dimensional space. Formally this constraint can be written as
\begin{equation}\nonumber
    \|f^{1}(x_{i}^{1})-f^{2}(x_{i}^{2})\|\leq \eta_{i}+\epsilon
\end{equation}
where $\eta_{i}$ is a variable that imposes consensus between the two views, and $\epsilon$ is a slack variable. In multi-view embedding, we conduct the embedding for multiple features simultaneously while considering the consistency and complement of different views. For example, the multi-view spectral embedding \citep{xia2010multiview} first builds a patch for a sample on each view, in which the arbitrary point and its $k$ nearest neighbors are forced to have similar outputs in the low-dimensional embedding space. Following this local consensus optimization, all the patches from different views are unified as a whole by global coordinate alignment. This can be seen as a global consensus optimization.

\subsection{Complementary Principle}

The complementary principle states that in a multi-view setting, each view of the data may contain some knowledge that other views do not have; therefore, multiple views can be employed to comprehensively and accurately describe the data. In machine learning problems involving multi-view data, the complementary information underlying multiple views can be exploited to improve the learning performance by utilizing the complementary principle.

\cite{Co-em} used the classifier learned on one view to label the unlabeled data, and then prepared these newly labeled examples for the next iteration of classifier training on another view. On the unlabeled data set, the models on two views therefore shared the complementary information with each other, which led to an improvement in the learning performance. \cite{Co-training-difference} studied why co-training style algorithms can succeed when there are no redundant views. They used different configurations of the same base learner, which can be seen as another kind of view, to describe the data in different approaches, and showed that when the diversity between the two learners is greater than the amount of errors, the performance of the learners can be improved by co-training style algorithms. The two classifiers which have different biases will label some examples with different labels. If the examples labeled by the classifier $h_{1}$ on one view are to be useful for the classifier $h_{2}$ on the other view, $h_{1}$  should contain some information that $h_{2}$ does not know. The two classifiers will thus exchange complementary information with each other and learn from each other under the complementary principle. As the co-training process proceeds, the two classifiers will become increasingly similar, until the performance cannot be further improved.

In multiple kernel learning, different kernels may correspond to different notations of similarity. Since different ways of measuring the similarity of the data have specific advantages, we resort to a learning method that makes an appropriate combination under the complementary principle rather than trying to establish which kernel works best. Thus all kinds of notations of similarity will work together to achieve an accurate evaluation of the data. In addition, different kernels can also use inputs from a variety views, possibly from alternative sources or modalities. Thus by considering the complementary information underlying various views of the data and combining multiple kernels from these distinct views, a comprehensive measurement of the similarity can be obtained.

One traditional solution for the multi-view problem is to concatenate vectors from different views into a new vector and then apply single-view learning algorithms straightforwardly on the concatenated vector. However, this concatenation causes over-fitting on a small training sample, and the specific statistical property of each view is ignored. For many applications with long feature vectors on more than one view as input, it is therefore reasonable to construct a shared low-dimensional representation for these views. In human pose inference, the image features and 3D poses can be seen as two complementary views that describe human poses. Several methods \citep{sgplvm, skie} have been designed to tackle this problem by constructing a latent subspace shared by multiple views, in which distinct views are connected with one another in this subspace, integrating the complementary information underlying different views. At inference, given a new observation on one view, it is possible to find the corresponding latent embedding, which is also connected with the point on the other view. \cite{xia2010multiview} developed a new spectral embedding algorithm, namely, multi-view spectral embedding (MSE), which encodes multi-view features to achieve a physical meaningful embedding. \cite{yu2012semi} proposed a semi-supervised multi-view distance metric learning (SSM-DML) for cartoon character retrieval. Since various low-level features can be extracted to represent the image, each feature space will give one measurement of similarity of the data, so it is difficult to decide which measurement is the most suitable. By considering the complementary information underlying distinct views, advantage can be taken of metric learning to construct a shared latent subspace to precisely measure the dissimilarity between different examples.

Both complementary and consensus principles play important roles in multi-view learning. For example, in co-training style algorithms, \cite{dasgupta2002pac} have shown that by minimizing the disagreement rate of the two hypotheses on two views respectively, the error rate of each hypothesis can be minimized. On the other hand, \cite{Co-training-difference} established that the reason for the success of co-training style algorithms is the extent of the diversity between the two learners; in other words, it is the complementary information in distinct views that influences the performance of co-training style algorithms. In addressing the problem of multi-view learning, both the consensus and complementary principles should be kept in mind to take full advantage of multiple views.

\section{View Generation}
\label{sec:3}

The priority for multi-view learning is the acquisition of redundant views, which is also a major difference from single-view learning. Multiple view generation not only aims to obtain the views of different attributes, but also involves the problem of ensuring that the views sufficiently represent the data and satisfy the assumptions required for learning. In this section, we will illustrate how to construct multiple views and how to evaluate these views.

\subsection{View Construction}

In practice, objects can frequently be described from different points of view. One classic multi-view example is the web classification problem. Usually, a web document can be described by either the words occurring on the page or the words contained in the anchor text of links pointing to this page. In many cases, however, no natural multiple views are available because of certain limitations, so that only one view may be provided to represent the data. Since it is difficult to straightforwardly conduct multi-view learning on this single view, the preliminary work of multi-view learning concerns the construction of multiple views from this single view.

Generating different views corresponds to feature set partitioning, which generalizes the task of feature selection. Instead of providing a single representative set of features, feature set partitioning decomposes the original set of features into multiple disjoint subsets to construct each view. A simple way to convert from a single view to multiple views is to split the original feature set into different views at random, and there indeed a number of experiments in multi-view learning employing this trick \citep{brefeld2005multi, multi_view_clustering, co_em_svm}. However, there is no guarantee that a satisfactory result will be obtained using this approach. Therefore, subsetting the feature set in a way that adheres to the multi-view learning paradigm is not a trivial task,  and is dependent on both the chosen learner and the data domain.

The random subspace method (RSM) \citep{ho1998random}, as an example of a random sampling algorithm, incorporates the benefits of bootstrapping and aggregation. Unlike bagging the bootstraps training samples, RSM performs the bootstrapping in the feature space. This method relies on an autonomous, pseudo random procedure to select a small number of dimensions from a given feature space. This selection is made and a subspace is fixed by giving all points a constant value (zero) in the unselected dimensions, in each pass. For a given feature space of $n$ dimensions, there are $2^{n}$ such selections that can be constructed. All the subspaces can then be regarded as different views of the data. While most other methods suffer from the curse of dimensionality, this method takes advantage of the high dimensionality. \cite{tao2006asymmetric} employed the random subspace method to sample several small sets of features to reduce the discrepancy between the training data size and the feature vector length. Based on the sampled subspaces, multiple SVMs can be constructed and then be combined to obtain a more powerful classifier to solve the over-fitting problem.

\cite{diview} conducted a thorough investigation of view generation for hyperspectral image data. Considering the key issues: diversity, compatibility and accuracy, several strategies have been proposed to construct multiple views for hyperspectral data, as follows. 1) Clustering: these methods involve feature aggregation based on similarity metrics, with the goal of promoting diversity between views. 2) Random selection: in conjunction with feature space bagging, random selection can result in greater information exploration from the spectral feature space and can eliminate the impact of generating uninformative or corrupted views. 3) Uniform band slicing: uniform division of the data across the full spectral range creates views that contain bands separated by equal intervals, thus guaranteeing view sufficiency. The authors also proposed that increasing the number of views to increase diversity, or increasing randomness from the feature space to avoid insufficient or noisy views, further improves performance.

With respect to learning problems involving textual documents, \cite{matsubara2005multi} proposed a pre-processing approach to easily construct different views required by multi-view learning algorithms. By identifying terms as bag-of-words and using different numbers of words to constitute each term, different representations of one document for different views can be obtained. This is a simple yet effective approach to the construction of multiple views for textual documents, although it is difficult to apply to other domains. \cite{wang2011novel} developed a novel technique to reshape the original vector representation of a single view into multiple matrix representations. For instance, a vector $x=[a,b,c,d,e,f]^{T}$ can be reshaped into two different matrices:
\begin{displaymath}
\left(               
  \begin{array}{ccc}   
    a & c & e\\  
    b & d & f\\  
  \end{array}
\right)
\quad and \quad                
\left(               
  \begin{array}{ccc}   
    a & b & c\\  
    d & e & f\\  
  \end{array}
\right)^{T}.
\end{displaymath}
Different ways of reshaping the vector induce multiple matrix patterns with a variety of dimensional sizes of rows and columns. These matrixes can be regarded as multiple independent or weaker correlated views of the input data. Utilizing the matrix representation, the required memory can be saved, new implicit information is introduced through the new constraint in the structure, and then the performance of the classifiers learned will be improved,  compared to the vector representation.

\cite{feature_decomposition} suggested a novel feature decomposition algorithm called Pseudo Multi-view Co-training (PMC) to automatically divide the features of a single view dataset into two mutually exclusive subsets. Considering the linear classifier, $f(x)=\textbf{w}\textbf{x}+b$ given the weight vector $\textbf{w}$, the optimization can be written as
\begin{equation}\label{}
    \min_{\textbf{w}_{1},\textbf{w}_{2}} \quad \log(e^{\mathcal{L}(\textbf{w}_{1};L)}+e^{\mathcal{L}(\textbf{w}_{2};L)}),
\end{equation}
where $\textbf{w}_{1}$ and $\textbf{w}_{2}$ are weight vectors for two classifiers respectively, and $\mathcal{L}(\textbf{w};L)$ is the log-loss over the dataset $L$. To make sure that the two classifiers are trained on different views of the dataset, for each feature $i$, at least one of the two classifiers must have a zero weight in the $i$-th dimension. This constraint can be written as
\begin{equation}\label{}
    \forall i, 1\leq i \leq d, \quad \textbf{w}_{1}^{i}\textbf{w}_{2}^{i}=0.
\end{equation}
In each iteration, solving the above optimization problem will automatically find an optimal split of the features.

To obtain the feature subsets automatically, \cite{sun2011view} turned to genetic algorithms (GAs) for help. Each bit in the binary bit strings in GAs is associated with one feature. If the $i$-th feature is selected, the $i$-th bit is 1, otherwise this bit is 0. Suppose the size of the population is $n$, then in each iteration, the best $n$ individuals will be selected as the next generation. Each individual in the final genetic population corresponds to a candidate feature subset, which can be regarded as one view of the data.

The literature shows that several kernel functions have been successfully used, such as the linear kernel, the polynomial kernel, and the Gaussian kernel. Since different kinds of kernel function correspond to different notations of similarity, it is reasonable instead of selecting one specific kernel function to describe the data to obtain an optimal combination of these kernel functions. These different kinds of kernel function can be seen as distinct views of the data, and the problem of how to learn the kernel combination can therefore be cast as multiple kernel learning.

The above view construction methods can be analyzed and categorized into three classes. The first class includes techniques that construct multiple views from meta data through random approaches. The second class consists of algorithms that reshape or decompose the original single-view feature into multiple views, such as the above matrix representations or different kernel functions. The third class is composed of methods that perform feature set partitioning automatically, such as PMC \citep{feature_decomposition}. This last type of algorithm bears some connections with the mature feature selection algorithms \citep{jain1997feature, guyon2003introduction}; however, there are significant differences between multi-view feature selection and single-view feature selection. In multi-view feature selection, the relationships between multiple views should additionally be considered, besides the information within each view.

\subsection{View Evaluation}

Constructing multiple views is just one task of view generation; another significant aspect is to evaluate these views and ensure their effectiveness for multi-view learning algorithms. Several approaches have been proposed in the multi-view learning literature that analyze the relationships between multiple views or cope with the problems resulting from the violation of view assumptions or the noise in the views.

\cite{muslea2002adaptive} first introduced a view validation algorithm which predicts whether or not the views are sufficiently compatible for solving multi-view learning tasks. This algorithm tries to learn a decision tree in a supervised way to discriminate between learning tasks according to whether or not the views are sufficiently compatible for multi-view learning. A set of features is designed to indicate how incompatible the views are, and the label of each instance is generated automatically by comparing the accuracy of single- and multi-view algorithms on a test set.

The assumption of view sufficiency does not generally hold in practice. For example, in the task of video concept detection, one frame contains an airplane and the other contains an eagle, but both frames may have the same color histogram feature. Therefore, it is difficult for the low-level visual features alone to sufficiently represent the concepts. \cite{yan2005semi} proposed semi-supervised cross-feature learning (SCFL) to alleviate the problems of co-training when some views are inadequate for learning concepts by themselves. When view sufficiency assumption fails, the main concern in applying co-training is that the additional training data associated with classification noise are likely to corrupt the initial classifiers. After labeling unlabeled data using the initial classifiers of two views, two separate classifiers from each view, based solely on the unlabeled data, are trained to eliminate this problem. With the assistance of validation data $V$, all four classifiers can be weighted combined to detect how much benefit can be achieved from the unlabeled data without hurting the performance of the initial classifiers. If the predictions from the unlabeled data are too noisy to use, the combined weights of the two classifiers newly learned on the unlabeled data can simply be zeroed, and we back off to the initial classifiers trained on the labeled data.

The performance of multi-view learning algorithms may be influenced by the noises in the views. \cite{christoudias2008multi} defined a view disagreement problem, stating that the samples from each view do not always belong to the same class but sometimes belong to an additional background class as a result of noise. To detect and filter view disagreement, a conditional view entropy $H(x^{i}|x^{j})$ was defined as a measure of the uncertainty in view $x^i$ given the observed view $x^{i}$. The conditional view entropy is expected to be larger when conditioning takes place on the background rather than the foreground. By thresholding the conditional view entropy, the samples whose views display disagreement can be discarded in each iteration of the co-training algorithm,  and then the performance of the classifiers is improved.

\cite{yu2011bayesian} proposed a probabilistic approach to co-training, called Bayesian co-training, which copes with per-view noise. This algorithm employs a latent variable $f_{j}$ for each view and a consensus latent variable, $f_{c}$ to model the agreement on different views. Finally $\psi(f_{j},f_{c})$ is defined to denote the compatibility between the $j$-th view and the consensus function and can be written as, $\psi(f_{j},f_{c})=exp(-\frac{\|f_{j}-f_{c}\|^{2}}{2\sigma_{j}^{2}})$. The parameters $\{\sigma_{j}\}$ act as reliability indicators and control the strength of interaction between the $j$-th view and the consensus latent variable. A small value of $\sigma_{j}$ has a strong influence on the view in the final output, whereas a large value allows the model to discount observations from that view. Thus the Bayesian co-training model can handle per-view noise, where each sample of a given view is assumed to be corrupted by the same amount of noise. \cite{christoudias2009co} extended Bayesian co-training to the heteroscedastic case, in which each observation can be corrupted by a different noise level. Assume that the latent functions can be corrupted with arbitrary Gaussian noise such that
\begin{equation}\label{}\nonumber
    \psi(f_{j},f_{c}) = \mathcal{N}(f_{j},\textbf{A}_{j}),
\end{equation}
where $\textbf{A}_{j}$ is the noise covariance matrix. When assuming i.i.d. noise, the noise matrix can be written as
\begin{equation}\label{}\nonumber
    \textbf{A}_{j}=diag(\sigma_{1,j}^{2},\cdots,\sigma_{N,j}^{2}),
\end{equation}
where $\sigma_{i,j}^{2}$ is the estimation of the noise corrupting sample $i$ in view $j$. Thus the heteroscedastic Bayesian co-training model can incorporate sample-dependent noise modeled by the per-view noise covariance matrices $\textbf{A}_{j}$.

In multiple kernel learning, different kernels may use inputs coming from various representations, possibly from a range of modalities or sources. These representations may have contrasting measures of similarity corresponding to different kernels, and can be regarded as different views of the data. In this case, combining kernels is one possible way to combine multiple information sources; however, in the real world, the sources may be corrupted by disparate noises, so when some of the kernels are noisy or irrelevant, it is necessary to optimize the kernel weights in the learning process. \cite{lewis2006support} compared the performances of unweighted and weighted sums of kernels on a gene functional classification task. They considered a case in which additional, noisy kernels are added to the system. As more noise is added to the system, the performance of the unweighted average deteriorates, but the weighted kernel approach learns to down-weight the noise kernels and hence continues to work well. Most multiple kernel learning algorithms are global techniques under the assumption of a per-view kernel weighting, and these methods therefore cannot cope with the presence of complex noise processes, such as heteroscedastic noise, or missing data. \cite{christoudias2009bayesian} presented a Bayesian localized approach for combining different feature representations with Gaussian processes that learns a local weighting over each view. Let $\overline{\textbf{X}}=[\textbf{X}^{1},\cdots,\textbf{X}^{V}]$ be the set of all observations with $V$ views, let $\textbf{Y}=[\textbf{y}_{1},\cdots,\textbf{y}_{N}]^{T}$ be the set of labels, and let $\textbf{f}=[\textbf{f}_{1},\cdots,\textbf{f}_{N}]^{T}$ be a set of latent functions. The Gaussian Process (GP) prior over the latent functions can be written as $p(\textbf{f}|\overline{\textbf{X}})=\mathcal{N}(0,\overline{\textbf{K}})$. If a Gaussian noise model is used, then $p(\textbf{Y}|\textbf{f})=\mathcal{N}(\textbf{f},\sigma^{2}\textbf{I})$ is obtained. The covariance function can be obtained by combining the covariances of feature representations in a non-linear manner; thus, classification is performed using the standard GP approach with common covariance function.

\cite{liu2011boosted} introduced two new confidence measures, namely, inter-view confidence and intra-view confidence, to describe the view sufficiency and view dependency issues in multi-view learning. Considering the sample $X$ associated with $M$ views, the observed data are represented as $X^{1},\cdots,X^{M}$ respectively; based on the mutual information definition, the inter-view confidence of $X$ is defined as
\begin{equation}\label{}\nonumber
    C_{inter}(X)=\sum_{i=1}^{M}\sum_{j=i}^{M}\frac{1}{I(X^{i},X^{j})},
\end{equation}
where $I(X^{i},X^{j})$ measures the mutual information between $X^{i}$ and $X^{j}$. By maximizing the inter-view confidence, the total data dependency is minimized. In addition, the authors proposed the calculation and minimization of the total inconsistency of labeled and unlabeled data iteratively in a semi-supervised manner. Thus the view sufficiency can be defined as
\begin{equation}\label{}\nonumber
    C_{intra}(X)=\sum_{i=1}^{M}\frac{1}{F(X_{i}^{L},X_{i}^{U},S_{i})},
\end{equation}
where $X_{i}^{L}$ and $X_{i}^{U}$ are the labeled and unlabeled dataset respectively, $S_{i}$ is the similarity matrix for view i, and F measures the data consistency between $X_{i}^{L}$ and $X_{i}^{U}$.

Correlation between views is an important consideration in subspace-based approaches for multi-view learning. \cite{hotelling1936relations} introduced canonical correlation analysis (CCA) to describe the linear relation between two views which aims to compute a low-dimensional shared embedding of both views of variables such that the correlations among the variables between the two views is maximized in the embedded space. Since the new subspace is simply a linear system of the original space, CCA can only be used to describe linear relation. Under Gaussian assumption, the CCA can also be used to test stochastic independence between two views of variables. \cite{akaho2006kernel} studied a hybrid approach of CCA with a kernel machine, called kernel canonical correlation analysis (KCCA), to identify non-linearly correlated projections between the two views. Formally, for two views $X\in \mathbb{R}^{d\times n}$ and $Y\in \mathbb{R}^{k\times n}$, CCA computes two projection vectors, $w_{x}\in \mathbb{R}^{d}$ and $w_{y}\in \mathbb{R}^{k}$, such that the following correlation coefficient:
\begin{equation}\nonumber
    \rho=\frac{w_{x}^{T}XY^{T}w_{y}}{\sqrt{(w_{x}^{T}XX^{T}w_{x})(w_{y}^{T}YY^{Y}w_{y})}}
\end{equation}
is maximized. Similarly in KCCA, we express the projection direction as $w_{x}=X\alpha$ and $w_{y}=Y\beta$, where $\alpha$ and $\beta$ are vectors of size $N$. Irrespective of whether CCA or KCCA is used, a sequence of correlation coefficients $\{\rho_{1},\rho_{2},\cdots\}$ arranged in descending order can be obtained. Several measures of association in the literature are constructed as functions of the correlation coefficients, of which the two most common association measures are as follows. One is the maximal correlation
\begin{equation}\label{}\nonumber
    r(X,Y)=\rho_{1},
\end{equation}
and the other is
\begin{equation}\label{}\nonumber
    r(X,Y)=-\sum_{i=1}\log(1-\rho_{i}^{2}).
\end{equation}

\section{View Combination}
\label{sec:4}

One traditional way to combine multiple views is to concatenate all multiple views into a single view to adapt to the single-view learning setting. However, this concatenation causes over-fitting on a small training sample and is not physically meaningful because each view has a specific statistical property. Thus we resort to advanced methods of combining multiple views to achieve the improvement in learning performance compared to single-view learning algorithms.

\begin{figure}[t]
\begin{center}
\includegraphics[width=\textwidth]{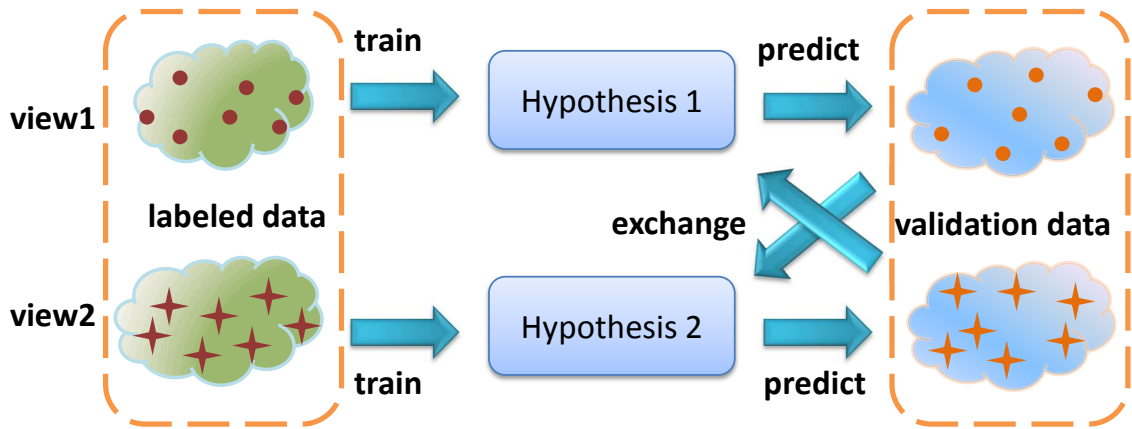}
   \caption{The process of co-training style algorithms.}\vspace{-5mm}
   \label{fig:2}
\end{center}
\end{figure}

Co-training style algorithms usually train separate but correlated learners on each view, and the outputs of learners are forced to be similar on the same validation points, as shown in Figure \ref{fig:2}, Under the consensus principle, the goal of each iteration is to maximize the consistency of two learners on the validation set. Certainly there may be some disagreement between the predictions from the two learners on the validation set; however, this disagreement is propagated back to the training set to help to train more accurate learners, thus minimizing the disagreement on the validation set in the next iteration.

Co-training is a classical algorithm in semi-supervised learning. In co-training, a classifier is trained on per-view, which only uses the features from that view. By maximizing the agreement on the predictions of two classifiers on the labeled dataset, as well as minimizing the disagreement on the predictions of two classifiers on the unlabeled dataset, the classifiers learn from each other and reach an optimal solution. Here, the unlabeled set is considered to be the validation set. In each iteration, the learner on one view labels unlabeled data which are then added to the training pool of the other learner; therefore, the information underlying two views can be exchanged in this scheme. Co-regularization can be regarded as a regularized version of the co-training algorithm. Unlike co-training, the co-regularization algorithm formally measures the agreement on two distinct views using Eq. \ref{Co-regularization}. By solving the corresponding objective problem, two optimal classifiers can be obtained.

If a validation set is not provided, for example in an unsupervised learning setting, it is necessary to train the classifier on each view as well as validate the combination of views on the same training set. \cite{Co-training-clustering} applied the idea of co-training to the unsupervised learning setting and proposed a spectral clustering algorithm for multi-view data. Under the assumption that the true underlying clustering would assign corresponding points in each view to the same cluster, this algorithm solves spectral clustering on individual graphs to obtain the discriminative eigenvectors $\textbf{U}_{1}(\textbf{U}_{2})$  in each view, then clusters points using $\textbf{U}_{1}(\textbf{U}_{2})$ and uses this clustering to modify the graph structure in views 2(1) respectively. This process is repeated for a number of iterations. Similar to many other multi-view clustering algorithms \citep{kumar2010co,Co-regularization-clustering}, multiple views in this setting are usually combined on the training set considering the consensus principle. In multi-view supervised learning problems, an implicit validation set is also employed to combine multiple views. For example, in the Bayesian co-training proposed by \cite{yu2011bayesian}, a Bayesian undirected graphical model for co-training through gauss process is constructed. A latent function $f_{c}$ is introduced to ensure the conditional independence between the output $y$ of each example and latent functions $f_{j}$ for each view. Thus $\{f_{c}\}$ can be seen as an implicit validation set which connects multiple views in a latent space.

\begin{figure}[t]
\begin{center}
\includegraphics[width=\textwidth]{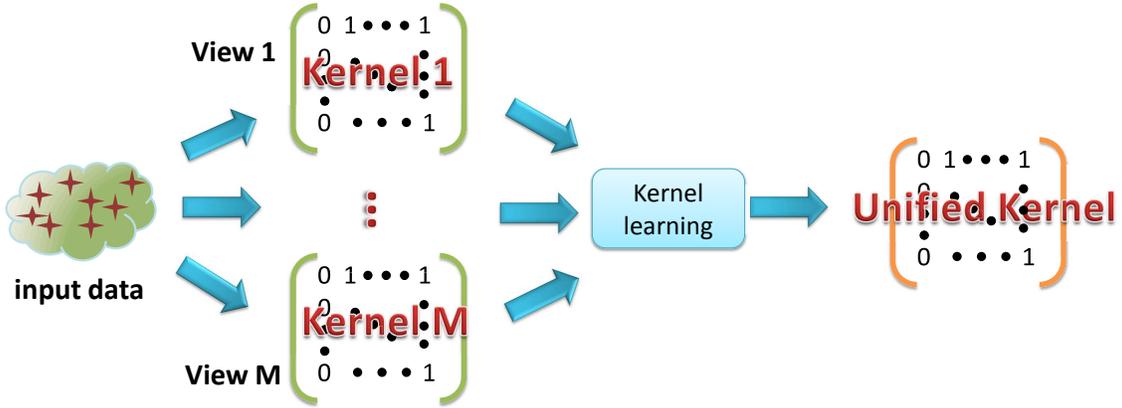}
   \caption{Sketch map of multiple kernel learning.}\vspace{-5mm}
   \label{fig:3}
\end{center}
\end{figure}

Instead of choosing a single kernel function for multiple kernel learning, it is better to use a set and allow an algorithm to choose suitable kernels and the kernel combination. Since different kernels may correspond to various notions of similarity or inputs coming from different representations, possibly from a number of sources or modalities, combining kernels is one possible way to integrate multiple information sources and find a better solution, as shown in Figure \ref{fig:3}. There are several ways in which the combination can be made, and each has its own combination parameter characteristics. These methods can be grouped into two categories: \begin{enumerate}
\item \textit{Linear combination methods}

There are several linear ways to combine multiple kernels. These methods have two basic categories:
\begin{itemize}
\item Direct summation kernel
\begin{equation}\label{}
    K(x_{i},x_{j})=\sum_{k=1}^{M}K_{k}(x_{i},x_{j}),
\end{equation}
\item Weighted summation kernel
\begin{equation}\label{}
    K(x_{i},x_{j})=\sum_{k=1}^{M}d_{k}K_{k}(x_{i},x_{j}).
\end{equation}
\end{itemize}

Using an unweighted sum gives equal preference to all kernels, which may not be ideal; a weighted sum may be a better choice. In the latter case, versions of this approach differ in the way they place restrictions on the kernel weights $\{d_{k}\}_{k=1}^{M}$. \cite{lanckriet2002learning, lanckriet2004learning} used a direct approach to optimize the unrestricted kernel combination weights. The combined kernel matrix is selected from the following set:
\begin{equation}\label{}
\nonumber
    \mathcal{K} = \{K:K=\sum_{k=1}^{M}d_{k}K_{k},K\geq 0, tr(K)\leq c\}.
\end{equation}
\cite{lanckriet2004learning} restricted the combination weights to non-negative values by selecting the combined kernel matrix from the set:
\begin{equation}\label{}
\nonumber
    \mathcal{K} = \{K:K=\sum_{k=1}^{M}d_{k}K_{k},d_{k}\geq 0, K\geq 0, tr(K)\leq c\}
\end{equation}
\cite{Joachims2001} followed the constraints $d_{k}\geq 0, \sum_{k=1}^{M}d_{k}=1$, and considered the convex combination of kernel weights. If only binary $d_{k}$ for kernel selection is allowed, the kernels whose  $d_{k}=0$ can be discarded and only the kernels whose $d_{k}=1$ are used. \cite{xu2009non} used this definition to perform feature selection. Usually the same weight is assigned to a kernel over the whole input space, which ignores the data distribution of each local region. \cite{gonen2008localized} proposed to assign different weights to kernel functions according to data distribution, and defined the locally combined kernel matrix as
\begin{equation}\label{}
    K(x_{i},x_{j})=\sum_{k=1}^{M}d_{k}(x_{i})K_{k}(x_{i},x_{j})d_{k}(x_{j}),
\end{equation}
where $d_{k}(x)$ is the gating function which chooses feature space as a function of input $x$.

\item \textit{Nonlinear combination methods}

Linear combinations of base kernels are limited, thus far richer representation can be achieved by combining kernels in other fashions. \cite{varma2009more} tried to use the products of base kernels and other combinations which yield positive definite kernels to perform multiple kernel learning; for example, the exponentiation and power way of combining kernels:
\begin{equation}\label{}
\nonumber
    K(x_{i},x_{j}) = exp(-\sum_{k=1}^{M}d_{k}x_{i}^{T}\textbf{A}_{k}x_{j}),
\end{equation}
or
\begin{equation}\label{}
\nonumber
    K(x_{i},x_{j}) = (d_{0}+\sum_{k=1}^{M}d_{k}x_{i}^{T}\textbf{A}_{k}x_{j})^{n}.
\end{equation}

Another work by \cite{cortes2009learning} is a non-linear kernel combination method based on kernel regression and the polynomial combination of kernels. They proposed to combine kernels as follows:
\begin{equation}\label{}
\nonumber
    K=\sum_{0\leq k_{1}+\cdots+k_{M}\leq d,k_{m}\geq 0}\mu_{k_{1}\cdots k_{M}}\prod_{m=1}^{M}K_{m}(x_{i},x_{j})^{k_{m}}，
\end{equation}
with
\begin{equation}\label{}
\nonumber
\mu_{k_{1}\cdots k_{M}}\geq 0.
\end{equation}
A special case is considered:
\begin{equation}\label{}
\nonumber
    K=\sum_{k_{1}+\cdots+k_{M}=d,k_{m}\geq 0}\prod_{m=1}^{M}\mu_{m}^{k_{m}}K_{m}(x_{i},x_{j})^{k_{m}},
\end{equation}
with
\begin{equation}\label{}
\nonumber
\mu_{k_{1}\cdots k_{M}}\geq 0.
\end{equation}
Consequently, the objective of the algorithm is to find the vector μ$\mu=(\mu_{1},\cdots,\mu_{M})^{T}$. However, the empirical results do not show consistent performance improvement, bringing into question whether the non-linear combination of kernel functions is necessary or efficient.
\end{enumerate}

\begin{figure}[t]
\begin{center}
\includegraphics[width=\textwidth]{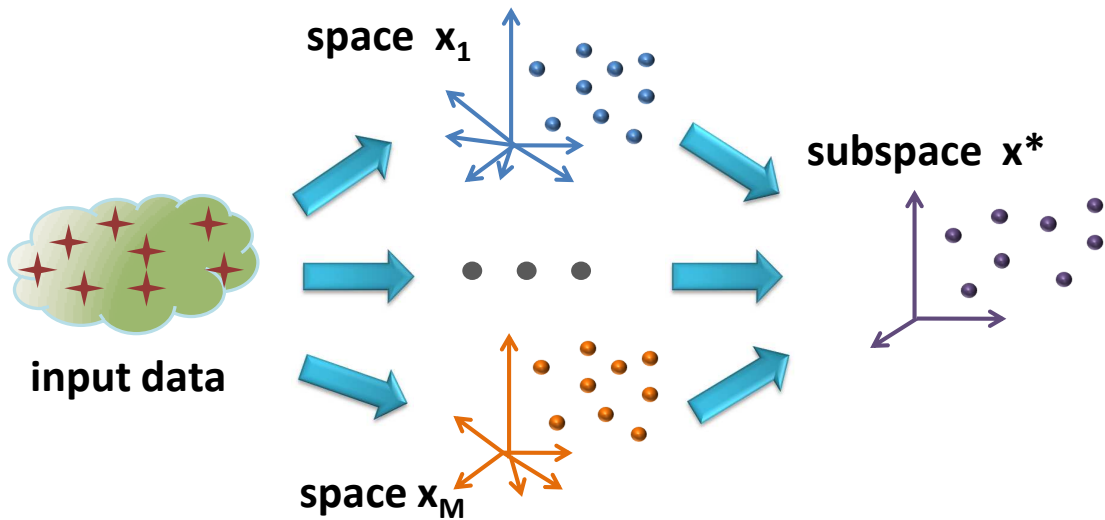}
   \caption{Sketch map of subspace learning for multi-view data.}\vspace{-5mm}
   \label{fig:4}
\end{center}
\end{figure}

Subspace learning-based approaches aim to obtain a latent subspace shared by multiple views by assuming that the input views are generated from this latent subspace, as illustrated in Figure \ref{fig:4}. In the literature on single-view learning, principal component analysis (PCA) is the time-honoured and simplest technique to exploit the subspace for single-view data. Canonical correlation analysis (CCA) can be viewed as the multi-view version of PCA, and it has became a general tool for performing subspace learning for multi-view data. Through maximizing the correlation between the two views in the subspace, CCA outputs one optimal projection on each view; however, since the subspace constructed by CCA is linear, it is impossible to straightforwardly apply this to many real world datasets exhibiting non-linearities. Thus the kernel variant of CCA, namely KCCA, can be thought of in terms of first mapping each data point to a higher space in which linear CCA operates. Both CCA and KCCA exploit the subspace in an unsupervised way, so that the label information is ignored. Motivated by the generation of CCA from PCA, multi-view Fisher discriminant analysis is developed to find informative projections with label information. \cite{gplvm} cast the Gaussian process as a tool to construct a latent variable model which could accomplish the task of non-linear dimensional reduction. \cite{multi_view_mn} developed a statistical framework that learns a predictive subspace shared by multiple views based on a generic multi-view latent space Markov network. \cite{quadrianto2011learning} studied the metric learning problem in cross-media retrieval tasks. The goal of metric learning for multi-view data is to learn metrics with which the original multi-view higher dimensional features can be projected into a shared feature space, so that the Euclidean distance in this space is meaningful not only within a single view, but also among different views. Since the subspace constructed through different methods usually has lower dimensionality than that of any input view, the ``curse of dimensionality'' problem is effectively eliminated, and given the subspaces, it is straightforward to conduct subsequent tasks such as classification and clustering.

After analyzing the various approaches above to combine multiple views, we sum up their similarities and differences as follows. (a) Co-training style algorithms usually train separate learners on distinct views, which are then forced to be consistent across views. Thus this kind of approach can be regarded as a late combination of multiple views because the views are considered independently while training base learners. (b) Multiple kernel learning algorithms calculate separate kernels on each view which are combined with a kernel-based method. This kind of approach can be thought of as an intermediate combination of multiple views because kernels (views) are combined just before or during the training of the learner. (c) Subspace learning-based approaches aim to obtain an appropriate subspace by assuming that input views are generated from a latent view. This kind of approach can be seen as the prior combination of multiple views because multiple views are straightforwardly considered together to exploit the shared subspace.

\section{Co-training Style Algorithms}
\label{sec:5}

Co-training \citep{Co-training} was one of the earliest schemes for multi-view learning. Since then, many variants have been developed. Besides the research on designing various algorithms, there are also a number of works on assumptions for co-training, which ensure the success of algorithms.

\subsection{Assumptions for Co-training}

Co-training considers a setting in which each example can be partitioned into two distinct views, and makes three main assumptions: (a) \emph{Sufficiency}: each view is sufficient for classification on its own, (b) \emph{Compatibility}: the target functions in both views predict the same labels for co-occurring features with high probability, and (c) \emph{Conditional independence}: the views are conditionally independent given the class label. The conditional independence assumption plays a critical role, but it is usually too strong to be satisfied in practice and several weaker alternatives have thus been considered.

\subsubsection{Conditional Independence Assumption}

\cite{Co-training} proved that when two sufficient views are conditionally independent given the class label, co-training can be successful. They gave a theorem that if the concept classes $C_{1,2}$ on view $X_{1,2}$ are learnable in the PAC model in spite of the classification noise, and if the conditional independence assumption is satisfied, then $(C_{1},C_{2})$ is learnable in the co-training model from unlabeled data only, given an initial weakly-useful predictor $h(x_{1})$. Specifically, let classification noise $(\alpha,\beta)$ be a setting in which true positive examples are incorrectly labeled (independently) with probability $\alpha$, and the true negative examples are incorrectly labeled (independently) with probability $\beta$. Again define $f(x)$ as the target concept and $p=Pr_{D}(f(x)=1)$ as the probability that a random example from $D$ is positive. The sum of the two noise rates satisfies
\begin{equation}
    \alpha +\beta <1-\varepsilon^{2}/(p(1-p)),
\end{equation}
with the probability at most $1-4\varepsilon^{2}$.

The above inequation gives some degree of justification for the co-training restriction on rule-based bootstrapping. However, it does not provide a bound on generalization error as a function of empirically measurable quantities; hence, based on the same conditional independence assumption,\cite{dasgupta2002pac} gave the PAC style bounds for co-training. Let $S$ be an i.i.d sample consisting of individual samples $s_{1},\cdots,s_{m}$. A partial rule $h$ on a dataset $X$ is a mapping from $X$ to the label set $\{1,\cdots,k,\bot\}$, where $k$ is the number of class labels and $\bot$ denotes the partial rule $h$ gives no opinion. Then with probability at least $1-\delta$ over the choice of S we have the following for all pairs of rules $h_{1}$ and $h_{2}$: if γ$\gamma_{i}(h_{1},h_{2},\delta/2)>0$ for $1\leq i \leq k$ then $f$ is a permutation and for all $1\leq i \leq k$,
\begin{eqnarray}
\nonumber
    {P}(h_{1}={i}|f(y)=i,h_{1}\neq\perp) \leq  \frac{1}{\gamma_{i}(h_{1},h_{2},\delta)}(\epsilon_{i}(h_{1},h_{2},\delta)
    +\hat{P}(h_{1}\neq{i}|h_{2}=i,h_{1}\neq\perp)),
\end{eqnarray}
where
\begin{eqnarray}
\nonumber
\epsilon_{i}(h_{1},h_{2},\delta) = \sqrt{\frac{\ln{2}(|h_{1}|+|h_{2}|)+\ln{2}/\delta}{2|S(h_{2}=i,h_{1}\neq\perp)|}}
\end{eqnarray}
\begin{eqnarray}
\gamma_{i}(h_{1},h_{2},\delta)=\hat{P}(h_{1}={i}|h_{2}=i,h_{1}\neq\perp)
-\hat{P}(h_{1}\neq{i}|h_{2}=i,h_{1}\neq\perp)
-2\epsilon_{i}(h_{1},h_{2},\delta).
\end{eqnarray}

Note that if the sample size is sufficiently large (relative to $|h_{1}|$ and $|h_{2}|$) then $\epsilon_{i}(h_{1},h_{2},\delta)$ is near to zero. Also note that if $h_{1}$ and $h_{2}$ have near perfect agreement when neither is $\perp$ then $\gamma_{i}(h_{1},h_{2},\delta)$ is near one. The agreement between $h_{1}$ and $h_{2}$ upper bounds the error of $h_{1}$. The co-training algorithm therefore needs to maximize the agreement on unlabeled data between classifiers based on different views under the conditional dependence assumption to improve the accuracy of each hypothesis.

\subsubsection{Weak Dependence Assumption}

The above-mentioned conditional independence assumption is overly strong to be satisfied for the two views in real applications. \cite{bootstrapping} relaxed the assumptions and found that weak dependence alone can lead to successful co-training. Given the mapping function $Y=y$, the conditional dependence of opposing-view rules $h_{1}$ and $h_{2}$ is defined as
\begin{equation}\nonumber
    d_{y} = \frac{1}{2}\sum_{u,v}|Pr[h_{1}=v|Y=y,h_{2}=u]-Pr[h_{1}=v|Y=y]|.
\end{equation}
If $h_{1}$ and $h_{2}$ are conditionally independent, then $d_{y}=0$. The $h_{1}$ and $h_{2}$ satisfy weak rule dependence just in case:
\begin{equation}\nonumber
    d_{y} \leq p_{2}\frac{q_{1}-p_{1}}{2p_{1}q_{1}},
\end{equation}
where $p_{1}=\min_{u}Pr[h_{2}=u|Y=y],p_{2}=\min_{u}Pr[h_{1}=u|Y=y]$, and $q_{1}=1-p_{1}$.

\subsubsection{Expansion Assumption}

\cite{Co-training-expansion} proposed a much weaker ``expansion'' assumption on the underlying data distribution and proved that it was sufficient for iterative co-training to succeed given appropriately strong PAC-learning algorithms on each feature set. Assume that examples are drawn from a distribution $D$ over an instance space $X$. Let $X^{+}$ and  $X^{-}$ denote the positive and negative regions of $X$ respectively. For $S_{1}\subseteq X_{1}$ and $S_{2}\subseteq X_{2}$, let $\textbf{S}_{i}(i=1,2)$ denote the event that an example $\langle x_{1},x_{2}\rangle$ has $x_{i}\in S_{i}$. If we think of $S_{1}$ and $S_{2}$ as confident sets in each view, then $Pr(\textbf{S}_{1}\wedge \textbf{S}_{2})$ denotes the probability mass on examples for which we are confident about both views, and $Pr(\textbf{S}_{1}\oplus \textbf{S}_{2})$ denotes the probability mass on examples for which we are confident about just one view. Define that $D^{+}$ is expanding if for any $S_{1}\subseteq X_{1}^{+}$ and $S_{2}\subseteq X_{2}^{+}$,
\begin{equation}
Pr(\textbf{S}_{1}\oplus \textbf{S}_{2}) \geq \epsilon \; \min[Pr(\textbf{S}_{1}\wedge \textbf{S}_{2}),Pr(\overline{\textbf{S}}_{1}\wedge \overline{\textbf{S}}_{2})].
\end{equation}
Another slightly stronger kind of expansion called ``left-right expansion'' can be defined as below.
$D^{+}$ is right-expanding if for any $S_{1}\subseteq X_{1}^{+}$ and $S_{2}\subseteq X_{2}^{+}$,
\\if
\begin{displaymath}
   \quad Pr(\textbf{S}_{1})\leq 1/2 , Pr(\textbf{S}_{2}|\textbf{S}_{1})\geq 1-\epsilon,
\end{displaymath}
then
\begin{displaymath}
    Pr(\textbf{S}_{2})\geq (1+\epsilon)Pr(\textbf{S}_{1}).
\end{displaymath}
$D^{+}$ is left-expanding if above holds with indices 1 and 2 reversed.

It can clearly be seen that if $S_{i}$ is the confident set in $X^{+}_{i}$ and this set is small $(Pr(\textbf{S}_{i})\leq 1/2)$, a classifier, which learns from positive data on the conditional distribution that $S_{i}$ induces over $X_{3-i}(i=1,2)$, is trained until it has error $\leq \epsilon$ on that distribution. The definition implies that the confident set on $X_{3-i}$ will have noticeably larger probability than $S_{i}$, so it is clear why this is useful for co-training.

\subsubsection{Large Diversity Assumption}

\cite{goldman2000enhancing} used two different supervised learning algorithms, and \cite{tri-training} used two different parameter configurations of the same base learner for co-training style algorithms without redundant views, but neither of them had addressed the reasons of their successes. Afterwards, \cite{Co-training-difference} showed that when the diversity between the two learners is larger than their errors, the performance of the learner can be improved by co-training style algorithms. The difference $d(h_{i},h_{j})$ between the two classifiers $h_{i}$ and $h_{j}$ implies the different biases between them, and the two classifiers will label some instances with different labels. If the examples labeled by the classifier $h_{i}$ are to be useful for the classifier $h_{j}$, $h_{i}$ should know some information that $h_{j}$ does not know. In other words, $h_{i}$ and $h_{j}$ should have significant differences. As the co-training process proceeds, the two classifiers will become increasingly similar and the difference between them will become smaller as the two classifiers label more and more unlabeled instances for each other. The co-training process would therefore not improve performance further after a number of learning rounds.

\subsection{Co-training}

Co-training was originally proposed for the problem of semi-supervised learning, in which there is access to labeled as well as unlabeled data. It considers a setting in which each example can be partitioned into two distinct views, and makes three main assumptions for its success: sufficiency, compatibility, and conditional independence.

In the original co-training algorithm \citep{Co-training}, given a set $L$ of labeled examples and a set $U$ of unlabeled examples, the algorithm first creates a smaller pool $U^{'}$ containing $u$ unlabeled examples. It then iterates the following procedure. First, use $L$ to train two naive Bayes classifiers $h_{1}$ and $h_{2}$ on the view $x_{1}$ and $x_{2}$ respectively. Second, allow each of these two classifiers to examine the unlabeled set $U^{'}$ and add the $p$ examples it most confidently labels as positive, and $n$ examples it most confidently labels as negative to $L$, along with the labels assigned by the corresponding classifier. Finally, the pool $U^{'}$ is replenished by drawing $2p+2n$ examples from $U$ at random.

\subsection{Co-EM}

The intuition behind the co-training algorithm is that classifier $h_{1}$ adds examples to the labeled set that classifier $h_{2}$ will then be able to use for learning. If the conditional independence assumption holds, then on average each added example will be as informative as a random example and learning should progress, subject to adding many examples belonging to the wrong class. If the independence assumption is violated, then on average the added examples will be less informative and co-training may not be successful. Instead of committing labels for the unlabeled examples, we thus choose to run EM in each view and give unlabeled examples probabilistic labels that may change from one iteration to another. This is the principal idea of co-EM \citep{Co-em}.

Co-EM outperforms co-training for many problems, but it requires the algorithm to process probabilistically labeled training data and the classifier to output class probabilities. Hence, the co-EM algorithm has only been studied with naive Bayes as  the underlying learner, even though Support Vector Machine (SVM) is known to better fit the characteristics of many classification problems. By reformulating the SVM in a probabilistic way and estimating the labels of unlabeled data with probabilities, \cite{co_em_svm} successfully developed a co-EM version of SVM to close this gap.

\subsection{Co-regularization}

Suppose we have two hypothesis spaces, $H^{1}$ and $H^{2}$, each of which contains a predictor that well-approximates the target function. In the case of co-training, these two are defined over different representations, or ``views'', of the data, and trained alternately to maximize mutual agreement on unlabeled examples. More recently, several papers have formulated those intuitions as joint complexity regularization, or co-regularization \citep{Co-regularization, manifold_Coregularization}, between $H^{1}$ and $H^{2}$ which are taken to be Reproducing Kernel Hilbert Spaces (RKHSs) of functions defined on the input space $X$. Given a few labeled examples ${(x_{i},y_{i})}_{i\in L}$ and a collection of unlabeled data $\{x_{i}\}_{i\in U}$, co-regularization learns a prediction function,
\begin{equation}\label{Co-regularization-1}
    f_{*}(x)=\frac{1}{2}(f_{*}^{1}(x)+f_{*}^{2}(x)),
\end{equation}
where $f_{*}^{1}\in H^{1}$ and $f_{*}^{1}\in H^{2}$ are obtained by solving the following optimization problem,
\begin{eqnarray}
\nonumber
  (f_{*}^{1},f_{*}^{2})  =  \min_{f^{1}\in H^{1},f^{2}\in H^{2}}\gamma_{1}\|f^{1}\|^{2}_{H^{1}}+\gamma_{2}\|f^{2}\|^{2}_{H^{2}} +\mu\sum_{i\in U}[f^{1}(x_{i})-f^{2}(x_{i})]^{2}+\sum_{i\in L}V(y_{i},f(x_{i})).
\end{eqnarray}
In this objective function, the first two terms measure complexity by RKHS norms $\|\cdot\|^{2}_{H_{1}}$ and $\|\cdot\|^{2}_{H_{2}}$ in $H_{1}$ and $H_{2}$ respectively, the third term enforces agreement among predictors on unlabeled examples, and the final term evaluates the empirical loss of the mean function $f-(f^{1}+f^{2})/2$ on the labeled data with respect to a loss function $V(\cdot,\cdot)$. The real valued parameters γ$\gamma_{1}$, $\gamma_{2}$ and $\mu$ allow different tradeoffs between the regularization terms. $L$ and $U$ are index sets over labeled and unlabeled examples respectively.

\subsection{Co-regression}

Most studies on multi-view and semi-supervised learning focus on classification problems, and regression problems can also be solved in a similar way. For instance, \cite{Co-training-regression} developed a co-training style semi-supervised regression algorithm called CoREG. This algorithm employs two k-nearest neighbor (kNN) regressors, each of which labels the unlabeled data for the other during the learning process. For the sake of choosing the appropriate unlabeled examples to label, CoREG estimates the labeling confidence by consulting the influence of the labeling of unlabeled examples on the labeled examples. The final prediction is made by averaging the regression estimates generated by both regressors. Inspired by the co-regularization algorithm, \cite{Co-regularization-regression} proposed a co-regression algorithm. Formally given $M$ views, the training instances $\{X_{v}\}_{v=1}^{M}$ with labels $y(x)\in \mathbb{R}$, and a finite set of instances $Z\subseteq X$ for which the labels are unknown, we attempt to find $f_{1}:X\rightarrow \mathbb{R},\cdots,f_{M}:X\rightarrow \mathbb{R}$ that minimize
\begin{eqnarray}
\nonumber
  Q(f) = \sum^{M}_{v=1}\Big[\sum_{x\in X_{v}}V(y(x),f_{v}(x))+\nu\|f_{v}(\cdot)\|^{2}\Big]
    + \lambda\sum^{M}_{u,v=1}\sum_{z\in Z}V(f_{u}(z),f_{v}(z)),
\end{eqnarray}
where the norms measuring complexity are in the respective Hilbert spaces, $V(y(x),f_{v}(x))$ evaluates the losses between the predictors and target values of labeled examples, and $V(f_{u}(z),f_{v}(z))$ imposes the agreement among predictors on unlabeled examples.

\subsection{Co-clustering}

The co-training algorithm was originally designed for semi-supervised learning, but the idea of co-training can also be applied in unsupervised and supervised learning settings. Under the assumption that the true underlying clustering will assign corresponding points in each view to the same cluster, several clustering techniques have been developed using the multi-view approach. \cite{multi_view_clustering} studied a multi-view version of the most frequently used clustering approaches such as k-means, k-medoids, and EM. Taking k-means as an example: in each iteration, run k-means in one view, then interchange the partition information to another view and run k-means in the second view again. After termination, compute a consensus mean for each cluster and view, then assign each example to one distinct cluster that is determined through the closed concept vector. Considering that spectral clustering algorithms have good performance on arbitrary shaped clusters and a well-defined mathematical framework, some methods are designed to utilize the idea of co-training to conduct spectral clustering \citep{kumar2010co,Co-regularization-clustering,Co-training-clustering}. For example, \cite{Co-training-clustering} developed a multi-view spectral clustering algorithm which solves spectral clustering on individual graphs to obtain the discriminative eigenvectors $\textbf{U}_{1}(\textbf{U}_{2})$ in each view, then clusters points using $\textbf{U}_{1}(\textbf{U}_{2})$ and uses this clustering to modify the graph structure in views 2(1) respectively. This process is repeated for a number of iterations.

\subsection{Graph-based Co-training}

Most co-training style algorithms focus on how to minimize the disagreement between two classifiers in order to obtain satisfactory performance of multi-view learners, thus these methods can be seen as disagreement-based approaches. Graph-based methods for co-training also exist; for instance, \cite{yu:bayesian, yu2011bayesian} proposed a Bayesian undirected graphical model for co-training through Gaussian process (GP). Suppose we have $m$ different views of $n$ data examples $\{x_{i}\}$ with outputs $\{y_{i}\}$. Let $f_{j}$ denote the latent function for the $j$-th view, and let $f_{j}\sim GP(0,\kappa)$ be its GP prior in view $j$. A latent function $f_{c}$ is then introduced to ensure conditional independence between the output $y$ and the $m$ latent functions ${f_{j}}$ for the $m$ views. At the functional level, the output $y$ depends only on $f_{c}$, and latent functions ${f_{j}}$ depend on each other only via the consensus function $f_{c}$. That is, we have the joint probability:
\begin{equation}
    p(y,f_{c},f_{1},\cdots,f_{m})=\frac{1}{Z}\psi(y,f_{c})\prod_{j=1}^{m}\psi(f_{j},f_{c}).
\end{equation}
In the ground network with $n$ data examples, let $\textbf{f}_{c}=\{f_{c}(x_{i})\}_{i=1}^{n}$ and $\textbf{f}_{j}=\{f_{j}(x_{i}^{j})\}_{i=1}^{n}$. The graphical model leads to the following factorization:
\begin{equation}\label{p-graphical}
    p(\textbf{y},\textbf{f}_{c},\textbf{f}_{1},\cdots,\textbf{f}_{m})=\frac{1}{Z} \prod_{i}\psi(y_{i},f_{c}(x_{i}))\prod_{j=1}^{m}\psi(\textbf{f}_{j})\psi(\textbf{f}_{j},\textbf{f}_{c}).
\end{equation}
Here, the within-view potential $\psi(\textbf{f}_{j})$ specifies the dependency structure within each view $j$, and the consensus potential $\psi(\textbf{f}_{j},\textbf{f}_{c})$ describes how the latent function in each view is related to the consensus function $f_{c}$. Employing a GP prior for each of the views, we can define the following potentials:
\begin{equation}
    \psi(\textbf{f}_{j})=exp(-\frac{1}{2}\textbf{f}_{j}^{T}\textbf{K}_{j}^{-1}\textbf{f}_{j}),\quad
    \psi(\textbf{f}_{j},\textbf{f}_{c}) = exp\Big(-\frac{\|\textbf{f}_{j}-\textbf{f}_{c}\|^{2}}{2\sigma_{j}^{2}}\Big).
\end{equation}
Integrating all the $m$ latent functions in Eq. (\ref{p-graphical}), we get the co-training kernel for multi-view learning as
\begin{equation}
    \textbf{K}_{c}=\Big[\sum_{j}(\textbf{K}_{j}+\sigma_{j}^{2}\textbf{I})^{-1}\Big]^{-1}.
\end{equation}
This co-training kernel reveals a previously unclear insight into how the kernels from different views are combined in multi-view learning and allows us to solve GP classification simply.

\cite{Co-traning-label} treated the co-training process as a combinative propagation over two views and unified the graph- and disagreement-based semi-supervised learning into one framework. In one view, the labels can be propagated from the initial labeled examples to unlabeled examples, and these newly-labeled examples can be added into the other view. The other view can then propagate the labels of the initial labeled examples and these newly labeled examples to the remaining unlabeled instances. This process can be repeated until the stopping condition is met.

\subsection{Multi-learner Algorithms}

\cite{goldman2000enhancing} presented a new ``co-training'' strategy for using unlabeled data to improve the performance of standard supervised learning algorithms. Without assuming that both of the views are sufficient for perfect classification, the only requirement of this co-training strategy is that its hypothesis partitions the example space into a set of equivalence classes. Assume that $A$ and $B$ are two different supervised algorithms, $U$ are unlabeled data, $L$ are the original labeled data, $L_{A}$ are the data that $B$ labeled for $A$, and $L_{B}$ are the data $A$ labeled for $B$. At the start of each iteration, train $A$ on the labeled examples $L\bigcup L_{A}$ to obtain the hypothesis $H_{A}$. Similarly, train $B$ on $L\bigcup L_{B}$ to obtain $H_{B}$. Each algorithm considers each of its equivalence classes and decides which to use to label data from $U$ for the other algorithm. This co-training algorithm repeats until neither $L_{A}$ nor $L_{B}$ change during an iteration.

\cite{tri-training} proposed another co-training style semi-supervised algorithm called tri-training, which does not require that the instance space be described with sufficient and redundant views, nor does it put any constraints on the supervised algorithm, as do \cite{goldman2000enhancing}. Tri-training generates three classifiers from the original labeled example set which are then refined using unlabeled examples in the iterations. For each iteration, an unlabeled example is labeled for a classifier if the other two classifiers agree on the labeling, under certain conditions.

The performance of traditional SVM-based relevance feedback approaches is often poor when the number of labeled feedback samples is small, thus \cite{li2006multitraining} developed a new machine learning technique, namely multi-training SVM (MTSVM), to mitigate this problem. MTSVM combines the merits of the co-training technique and a random sampling method in the feature space. However,simply using the co-training algorithm with SVM is not realistic, because the co-training algorithm requires that the initial sub-classifiers have good generalization ability before the co-training procedure commences. Thus the authors employed classifier committee learning to enhance the generalization ability of each sub-classifier. Initially, a series of subsets of feature - in other words, multiple views of the data can be obtained from the original input feature using the random subspace method. Multiple classifiers can then be learned on these generated views and can train one another in a semi-supervised relevance feedback setting. Finally, the majority voting rule is used to generate the optimal classifier.

\section{Multiple Kernel Learning}
\label{sec:6}

Multiple Kernel Learning (MKL) was originally developed to control the search space capacity of possible kernel matrices to achieve good generation, but it has been widely applied to problems involving multi-view data. This is because kernels in MKL naturally correspond to different views and combining kernels appropriately may improve learning performance. \cite{gonen2011multiple} have reviewed the literature on MKL. Since MKL can be regarded as just one part of multi-view learning, we place more weight on the connections between MKL and those parts; in this section, we illustrate the representative MKL algorithms and theoretical studies to present a complete picture in this survey.

\subsection{Boosting Methods}

Inspired by ensemble and boosting methods \citep{duffy2000leveraging, friedman2001elements}, \cite{bennett2002mark} proposed the Multiple Additive Regression Kernels (MARK) algorithm which considers a large library of kernel matrices formed by different kernel functions and parameters. The decision function is modified as
\begin{equation}\label{}
    f(x)=\sum_{i=1}^{N}\sum_{k=1}^{M}d_{i}^{k}K_{k}(x_{i}^{m},x^{m})+b,
\end{equation}
which is composed of a linear combination of heterogeneous kernel functions $K_{1},\cdots,K_{M}$, and each kernel can be of any type; for example, $\{K_{k}\}$ could be RBF kernels with different parameters. Like ensemble methods, each column of the kernel is treated as a hypothesis and the kernel columns are generated on the fly. Gradient-based ensemble algorithms, such as gradient boosting, can be adapted to this optimization problem.

Column Generation (CG) techniques have been widely used for solving large scale linear programs (LPs). \cite{bi2004column} used the 2-norm regularization approach to extend LPBoost to a quadratic program (QP), so that many successful formulations, such as classic SVMs, ridge regression, etc, could benefit from CG techniques.

\cite{crammer2002kernel} used the boosting paradigm to perform the kernel construction process. Since numerous interpretations of AdaBoost and its variants regard the boosting process as a procedure that attempts to minimize classification error, the boosting methodology can be modified to work with kernels by rewriting the loss functions for a pair of examples $(x_{1},y_{1})$ and $(x_{2},y_{2})$ as
\begin{eqnarray}
\nonumber
  ExpLoss(K(x_{1},x_{2}),y_{1}y_{2}) &=& \exp(-y_{1}y_{2}K(x_{1},x_{2})) \\
  \nonumber
  LogLoss(K(x_{1},x_{2}),y_{1}y_{2}) &=& \log(1+exp(-y_{1}y_{2}K(x_{1},x_{2}))).
\end{eqnarray}
A pair of instances is viewed as a single example and pairs of the same labels are regarded as positively labeled examples, while pairs of opposite labels are seen as negatively labeled examples. Along similar lines to boost algorithms for classification, the combined kernel matrix can be updated iteratively using one of these two loss functions.

\subsection{Semi-Definite Programming}

The general form of Semi-Definite Programming (SDP) is
\begin{eqnarray}
  \min_{x} &\quad& c^{T}x \\
  \nonumber
   \texttt{s.t.}&\quad& F(x)=F_{0}+x_{1}F_{1}+\cdots+x_{n}F_{n}\geq 0  \\
   \nonumber
   &\quad& Ax=b,
\end{eqnarray}
where $x\in \mathbb{R}^{p}$ and $F_{i}=F_{i}^{T}\in \mathbb{R}^{p\times p}$. Note that the object is linear in the unknown $x$, and that both inequality and equality constraints are linear in $x$.

\cite{lanckriet2002learning, lanckriet2004learning} showed how the kernel matrix can be learned from data via SDP techniques. In particular, if all labels of data are known,  the task is to find the kernel matrix $K$ which is maximally aligned with the set of labels $y$, and then this problem is formulated as
\begin{eqnarray}\label{eq:sdp}
  \max_{A,K} &\quad& \langle K,yy^{T}\rangle \\\nonumber
  \texttt{s.t.} &\quad& trace(A)\leq 1 \\\nonumber
   &\quad& \left(
             \begin{array}{cc}
               A & K^{T} \\
               K & I \\
             \end{array}
           \right)\geq 0
    \\\nonumber
   &\quad& K \geq 0.
\end{eqnarray}
Given the labeled training set $S_{n_{tr}}=\{(x_{1},y_{1}),
\cdots,(x_{n_{tr}},y_{n_{tr}})\}$ and the unlabeled test set $T_{n_{t}}\{x_{n_{tr}+1},\cdots,x_{n_{tr}+n_{t}}\}$, formally, we consider a kernel matrix has the form:
\begin{equation}\label{}
    K=\left(
        \begin{array}{cc}
          K_{tr} & K_{trt} \\
          K_{trt}^{T} & K_{t} \\
        \end{array}
      \right),
\end{equation}
where $K_{ij}=\langle\phi(x_{i}),\phi(x_{j})\rangle$, $i,j=1,\cdots,n_{tr},n_{tr}+1,n_{tr}+n_{t}$. The goal is then to learn the optimal mixed block $K_{trt}$ and the optimal ``test data block'' $K_{t}$ by optimizing a cost function over the ``training data block'' $K_{tr}$. Under the constraint $K=\sum_{i-1}^{M}\mu_{i}K_{i}$, where the set $\mathcal{K}=\{K_{1},\cdots,K_{M}\}$ is given and $\mu_{i}$ are to be optimized, we can replace $K$ with $K_{tr}$ in Eq. (\ref{eq:sdp}) and obtain the SDP formulation for learning the kernel matrix.

\subsection{Quadratically Constrained Quadratic Program (QCQP)}

\cite{bach2004multiple} introduced a novel classification algorithm called support kernel machine (SKM). Given a decomposition of $\mathbb{R}^{k}$ as a product of $m$ blocks: $\mathbb{R}^{k}=\mathbb{R}^{k_{1}}\times\cdots \times \mathbb{R}^{k_{m}}$, then each data $x_{i}$ can be decomposed into $m$ block components, $x_{i}=\{x_{1i},\cdots,x_{mi}\}$. The aim is to find a linear classifier, $y=\texttt{sign}(w^{T}x+b)$, where $w={w_{1},\cdots,w_{m}}$. To obtain the sparsity of the vector $w$ and make most of the components in $w$ zero, the 1-norm and 2-norm are used to penalize $w$. Thus the primal problem can be formulated as follow:
\begin{eqnarray}
\min && \frac{1}{2}(\sum_{j=1}^{m}d_{j}\|w_{j}\|_{2})^{2}+C\sum_{i=1}^{n}\xi_{i} \\\nonumber
\texttt{w.r.t.}  && w\in \mathbb{R}^{k}=\mathbb{R}^{k_{1}}\times\cdots \times \mathbb{R}^{k_{m}},\; \xi\in \mathbb{R}^{n}_{+},\; b\in \mathbb{R} \\\nonumber
\texttt{s.t.} && y_{i}(\sum_{j}w_{j}^{T}x_{ji}+b)\geq 1-\xi_{i},\forall i\in\{1,\cdots,n\}.
\end{eqnarray}

This optimization problem can be seen as a second order cone program (SOCP) problem, and then the dual problem is given by:
\begin{eqnarray}\label{eq:skm_dual}
  \min && \frac{1}{2}\gamma^{2}-\alpha^{T}e \\\nonumber
  \texttt{w.r.t.} && \gamma \in \mathbb{R},\; \alpha\in\mathbb{R}^{n} \\\nonumber
  \texttt{s.t.} && 0\leq \alpha \leq C, \; \alpha^{T}y=0\\\nonumber
  &&\|\sum_{i}\alpha_{i}y_{i}x_{ji}\|_{2}\leq d_{j}\gamma, \forall j\in\{1,\cdots,m\},
\end{eqnarray}
which is exactly equivalent to the QCQP formulation of \cite{lanckriet2004learning}. However, the advantage of this SOCP formulation is that \cite{bach2004multiple} developed an SMO algorithm for the SKM with Moreau-Yosida regularization, and transformed the primal problem as:
\begin{eqnarray}
\min && \frac{1}{2}(\sum_{j=1}^{m}d_{j}\|w_{j}\|_{2})^{2}+\frac{1}{2}\sum_{j}a_{j}^{2}\|w_{j}\|_{2}^{2}
+C\sum_{i=1}^{n}\xi_{i} \\\nonumber
\texttt{w.r.t.}  && w\in \mathbb{R}^{k}=\mathbb{R}^{k_{1}}\times\cdots \times \mathbb{R}^{k_{m}},\; \xi\in \mathbb{R}^{n}_{+},\; b\in \mathbb{R} \\\nonumber
\texttt{s.t.} && y_{i}(\sum_{j}w_{j}^{T}x_{ji}+b)\geq 1-\xi_{i},\forall i\in\{1,\cdots,n\},
\end{eqnarray}
where $\{a_{j}\}$ are the MY-regularization parameters.

\subsection{Semi-infinite Linear Program (SILP)}

\cite{sonnenburg2006general, sonnenburg2006large} followed a different direction and formulated the problem as a semi-infinite linear program (SILP). Beginning with Eq. (\ref{eq:skm_dual}), the equivalent multiple kernel learning dual is modified as:
\begin{eqnarray}
  \min && \gamma \\\nonumber
  \texttt{w.r.t.} && \gamma \in \mathbb{R},\; \alpha\in\mathbb{R}^{n} \\\nonumber
  \texttt{s.t.} && 0\leq \alpha \leq C, \; \alpha^{T}y=0\\\nonumber
  \forall k && \underbrace{  \frac{1}{2}\sum_{i=1}^{N}\sum_{j=1}^{N}\alpha_{i}\alpha_{j}
  y_{i}y_{j}K_{k}(x_{i}^{k},x_{j}^{k})-\sum_{i=1}^{N}\alpha_{i}}_{S_{k}(\alpha)}\leq\gamma,
\end{eqnarray}
which may be solved by
\begin{equation}\label{eq:silp5}
    L=\gamma+\sum_{k=1}^{M}\beta_{k}(S_{k}(\alpha)-\gamma)
\end{equation}
minimized w.r.t $\alpha$ and maximized w.r.t $\beta$. Setting the derivative w.r.t. to $\gamma$ to zero, the constraint $\sum_{k}\beta_{k}=1$ is obtained. Eq. (\ref{eq:silp5}) can then be simplified to a min-max problem
\begin{eqnarray}
\max_{\beta} \; \min_{\alpha} && \sum_{k=1}^{M}\beta_{k}S_{k}(\alpha) \\\nonumber
  \texttt{s.t.} && \sum_{i=1} \alpha_{i}y_{i}=0 \; \sum_{k=1}^{M}\beta_{k}=1.
\end{eqnarray}
Assume that $\alpha^{*}$ is the optimal solution, and given the definition of $\theta=L=S(\alpha^{*},\beta)$, Eq. (\ref{eq:silp5}) is equivalent to the following SILP problem:
\begin{eqnarray}
  \max && \theta \\\nonumber
  \texttt{s.t.} &&
  \sum_{k}\beta_{k}=1,\; \sum_{k}\beta_{k}S_{k}(\alpha)\geq \theta \\\nonumber
  \forall \alpha && 0\leq \alpha \leq C, \; \sum_{i}y_{i}\alpha_{i}=0,
\end{eqnarray}
where $\theta$ and $\beta$ are only linearly constrained, but there are a large number of constraints due to the possible values of $\alpha$.

Compared to the SDP and QCQP, the SILP formulation has a lower computational complexity, and this SILP problem can be efficiently solved using an off-the-shelf LP solver and a standard SVM implementation. Thus it allows us to efficiently handle more than a hundred thousand examples or several hundred kernels.

\subsection{Simple MKL}

\cite{rakotomamonjy2007more, rakotomamonjy2008simplemkl} departed from the framework proposed by \cite{bach2004multiple} and presented a different primal problem for multiple kernel learning through an adaptive 2-norm regularization formulation. Inspired by the multiple smoothing splines framework \citep{wahba1990spline},  the proposed primal formulation is
\begin{eqnarray}
  \min && \sum_{k}\frac{1}{d_{k}}\|w_{k}\|^{2}+C\sum_{i}\xi_{i} \\\nonumber
  \texttt{s.t.} && y_{i}(\sum_{k}w_{k}x_{i}^{k}+b)\geq 1-\xi_{i} \\\nonumber
   && \sum_{k}d_{k}=1 \\\nonumber
   && \xi_{i}\geq 0,d_{k}\geq 0 \forall i, \forall k.
\end{eqnarray}
Note that the $d_{k}$ controls the smoothness of kernel function, and the 1-norm constraint on the vector $d$ will lead to a sparse decision function with few basis kernels. By Defining the optimal SVM objective value $J(d)$ as
\begin{eqnarray}
  \min && J(d)=\sum_{k}\frac{1}{d_{k}}\|w_{k}\|^{2}+C\sum_{i}\xi_{i}  \\\nonumber
  \texttt{s.t.} && y_{i}(\sum_{k}w_{k}x_{i}^{k}+b)\geq 1-\xi_{i} \\\nonumber
   && \xi_{i} \geq 0,
\end{eqnarray}
the primal optimization problem can then be reformulated as
\begin{equation}\label{}
    \min_{d} J(d) \; \texttt{s.t.} \sum_{k}d_{k}=1,d_{k}\geq 0.
\end{equation}

The overall procedure to solve this problem consists of two steps: first, solving a canonical SVM optimization problem $J(d)$ with given $d$; second, updating d by gradient descent while ensuring that the constraints on $d$ are satisfied. This novel multiple kernel learning framework is called simple MKL, which has been shown to be more efficient than the SILP problem.

\cite{chapelle2008second} investigated the use of second order optimization approaches to solve the MKL problem, and propose hessian MKL as an extension of simple MKL. In each iteration, hessian MKL updates the kernel weights using a Newton step found by minimizing a QP problem. The result shows that hessian MKL outperforms simple MKL in terms of computational efficiency.

The SILP approach often suffers from slow convergence because it updates kernel weights based only on the cutting plane model. The simple MKL is efficient; however, it does not use the gradients computed in previous iterations, which can be useful in improving the efficiency of the search. \cite{xu2009extended} extended the level method, and applied it to multiple kernel learning to overcome the drawbacks of SILP \citep{sonnenburg2006large} and simple MKL \citep{rakotomamonjy2007more}. Following the SILP method, this algorithm has an extra step to adjust the solution for kernel weights obtained from a cutting plan model, through a projection to a level set. This adjustment ensures the new solution is close to the current solution and reduces the objective function.

\subsection{Group-LASSO Approaches}

It is reasonable to consider the group structure between the combined kernels when the kernels can be partitioned into groups which correspond to subsets of inputs or sources. In the learning process, it is desirable to suppress the kernels or groups that are irrelevant for the classification task, otherwise all the kernels belonging to the same groups which are relevant to the task will be selected. Based on this idea, \cite{szafranski2008composite, szafranski2010composite} developed the Composite Kernel Learning (CKL) approach, which extends the multiple kernel learning problem to take into account the group structure among kernels and constructs the relationship with group-LASSO \citep{yuan2006model}. The MKL formulation of \cite{rakotomamonjy2008simplemkl} is modified to obtain the following formulation of CKL:
\begin{eqnarray}
  \min && \frac{1}{2}\sum_{k}\frac{1}{d_{k}}\|w_{k}\|^{2}+C\sum_{i}\xi_{i} \\\nonumber
  \texttt{s.t.} && y_{i}(\sum_{k}w_{k}x_{i}^{k}+b)\geq 1-\xi_{i} \\\nonumber
   && \xi_{i}\geq 0 \\\nonumber
   && \sum_{G}\left(\|G\|^{p}(\sum_{k\in G}d_{k}^{1/q})^{q}\right)^{1/(p+q)}\leq 1 \\\nonumber
   && d_{k}\geq 0,
\end{eqnarray}
where $p$ and $q$ are set according to the problem at hand, $G$ denotes one subset of kernels, and $\|G\|$  is the size of group $G$. Note the third constraint: in particular cases where $p=0,q=1$, a LASSO type penalty is imposed on the RKHS norms, and when $p=1,q=0$, a group-LASSO type penalty is imposed on the RKHS norms.

\cite{xu2010simple} discussed the connection between multiple kernel learning and the group-LASSO regularizer, and generalized MKL to $L_{p}$-MKL which constrains the $p$-norm kernel weights. This proposed algorithm provides a unified solution for the entire family of $L_{p}$ models, besides which the kernel weights can be calculated by a closed-form formulation without dependence on other commercial software. \cite{subrahmanya2010sparse} proposed an algorithm called Sparse Multiple Kernel Learning (SMKL), which generalizes group-feature selection to kernel selection by introducing a log-based penalty over the groups. This method can automatically select the optimal number of sources from a large candidate list with a sparser solution compared to the existing multiple kernel learning framework.

\subsection{Bounds for Learning Kernels}

The most common family of kernels examined in multiple kernel learning is that of non-negative or convex combination of some fixed kernels constrained by a trace condition, which can be viewed as an $L_{1}$ or $L_{2}$ regularization, or $L_{p}$ regularization with other values of $p$.

\cite{lanckriet2004learning} showed that when a kernel is chosen from a convex combination of $k$ base kernels, the estimation error of the learned classifier is bounded by $O(\sqrt{\frac{k/\gamma^{2}}{n}})$, where $\gamma$ is the margin of the learned classifier under the kernel. This bound converges and can be viewed as the first informative generalization bound for this family of kernels; however, the multiplicative interaction between the margin complexity term $1/\gamma^{2}$ and the number of base kernels $k$ does not encourage the use of too many base kernels. It suggests that learning even a few kernel parameters leads to a multiplicative increase in the required sample size. \cite{srebro2006learning} presented a generalization bound for a kernel family with pseudo-dimension of $d_{\phi}$. The pseudo-dimension of most kernel families is similar to our intuitive notion of the dimensionality of the family; in particular, the pseudo-dimension of a family of linear or convex combinations of $k$ base kernels is at most $k$. The estimation error for SVMs with margin $\gamma$ is bounded by $\sqrt{O(d_{\phi}+1/\gamma^{2})/n}$, which  establishes that the bound on the required sample size, $O(d_{\phi}+1/\gamma^{2})$ grows only additive with the dimensionality of the allowed kernel family. \cite{ying2009generalization} showed that the generalization analysis of the regularized kernel learning system reduces to investigation of the suprema of the Rademacher chaos process of order two over candidate kernels, and they used metric entropy integrals and the pseudo-dimension of the set of candidate kernels to estimate the empirical Rademacher chaos complexity. For a pseudo-dimension of $k$, as in the case of a convex combination of $k$ base kernels, their bound is in $O(\sqrt{k(R^{2}/\rho^{2})(\log(m)/m)})$ and is thus multiplicative in $k$. Based on a combinatorial analysis of the Rademacher complexity of the hypothesis set under consideration, \cite{cortes2010generalization} presented another generalization bound with an $L_{1}$ constraint that has only a logarithmic dependency on the kernel number $k$. The bound is in $O(\sqrt{\frac{(\log{k})R^{2}/\rho^{2}}{m}})$, thus it is valid for a very large number of kernels, in particular for $k\gg m$, and it contains only a $\sqrt{\log{k}}$ dependency on the number of kernels, which is tight and considerably more favorable. Assuming the different views corresponding to the different kernels to be uncorrelated, \cite{kloft2011local} derived an upper bound on the local Rademacher complexity of $L_{p}$-norm multiple kernel learning. Given the number of kernels $M$ and the radius $D$, the bound for centered identical independent kernels is of the order $O(\sqrt{\sum_{j}^{\infty}\min{(rM,D^{2}M^{\frac{2}{p^{*}}}\lambda_{j})}})$. From the upper bound, a tighter excess risk bound than previous approaches is obtained, which achieves a fast convergence rate of the order $O(n^{-\frac{\alpha }{1+\alpha}})$, where $\alpha$ is the minimum eigenvalue decay rate of the individual kernels.

\section{Subspace Learning-based Approaches}
\label{sec:7}

Subspace learning-based approaches aim to obtain a latent subspace shared by multiple views by assuming that the input views are generated from this subspace. Besides the well known canonical correlation analysis (CCA), other more effective methods to construct the subspaces have recently become available.

\subsection{Algorithms based on CCA}

Canonical correlation analysis (CCA) is a technique for modeling the relationships between two (or more) sets of variables, and it has been applied with great success on a variety of learning problems dealing with multi-view data.

\subsubsection{A review of CCA}

For $X\in \mathbb{R}^{D_{1}\times N}$ and $Y\in \mathbb{R}^{D_{2}\times N}$, CCA computes two projection vectors, $w_{x}\in \mathbb{R}^{D_{1}}$ and $w_{y}\in \mathbb{R}^{D_{2}}$, such that the following correlation coefficient:
\begin{equation}
    \rho=\frac{w_{x}^{T}XY^{T}w_{y}}{\sqrt{(w_{x}^{T}XX^{T}w_{x})(w_{y}^{T}YY^{Y}w_{y})}}
\end{equation}
is maximized. Since $\rho$ is invariant to the scaling of $w_{x}$ and $w_{y}$, CCA can be formulated equivalently as
\begin{eqnarray}\label{cca_2}
  \max_{w_{x},w_{y}} && w_{x}^{T}XY^{T}w_{y} \\
  \nonumber
  \texttt{s.t.} & & w_{x}^{T}XX^{T}w_{x}=1, \quad w_{y}^{T}YY^{T}w_{y}=1.
\end{eqnarray}
Assuming $YY^{T}$ is nonsingular, then $w_{x}$ can be obtained by solving the following optimization problem:
\begin{eqnarray}\label{cca_3}
  \max_{w_{x},w_{y}} && w_{x}^{T}XY^{T}(YY^{T})^{-1}YX^{T}w_{y} \\
  \nonumber
  \texttt{s.t.} & & w_{x}^{T}XX^{T}w_{x}=1
\end{eqnarray}

Both formulations in Eqs. (\ref{cca_2}) and (\ref{cca_3}) attempt to find the eigenvectors corresponding to the top eigenvalues of the following generalized eigenvalue problem:
\begin{equation}\label{cca_4}
    XY^{T}(YY^{T})^{-1}YX^{T}w_{x}=\eta XX^{T}w_{x},
\end{equation}
where $\eta$ is the eigenvalue corresponding to the eigenvector $w_{x}$.

\subsubsection{Kernel CCA}

Canonical correlation analysis (CCA) is a linear feature extraction algorithm, but for many real world datasets exhibiting non-linearities, it is impossible for a linear projection to capture the properties of the data. Kernel methods provide a way to deal with the non-linearities by mapping the data to a higher dimensional space and then applying linear methods in that space.

Formally given a pair of datasets $X\in \mathbb{R}^{D_{1}\times N}$  and $Y\in \mathbb{R}^{D_{2}\times N}$, CCA seeks to find linear projections $w_{x}\in \mathbb{R}^{D_{1}}$ and $w_{y}\in \mathbb{R}^{D_{2}}$ such that, after projecting, the corresponding examples in the two datasets are maximally correlated in the projected space. To obtain the kernel formulation of CCA, dual representation is engaged by expressing the projection direction as $w_{x}=X\alpha$ and $w_{y}=Y\beta$ where $\alpha$ and $\beta$ are vectors of size $N$. In the dual formulation, the correlation coefficient between $X$ and $Y$ can be written as:
\begin{equation}\label{kcca_1}
    \rho=\max_{\alpha,\beta}\frac{{\alpha}^{T}X^{T}XY^{T}Y\beta}{\sqrt{{\alpha}^{T}X^{T}XX^{T}X\alpha \times {\beta}^{T}Y^{T}YY^{T}Y\beta }}
\end{equation}

Now using the fact that $K_{x}=X^{T}X$ and $K_{y}=Y^{T}Y$ are the kernel matrices for $X$ and $Y$, kernel CCA amounts to solving the following problem:
\begin{eqnarray}\label{kcca_2}
  \max_{\alpha, \beta} &=& \frac{\alpha^{T}K_{x}K_{y}\beta}{\sqrt{\alpha^{T}K_{x}^{2}\alpha\times \beta^{T}K_{y}^{2}\beta}} \\
  \nonumber
  \texttt{s.t.} & & \alpha^{T}K_{x}^{2}\alpha=1,\quad \beta^{T}K_{y}^{2}\beta=1.
\end{eqnarray}

KCCA works by using the kernel matrices $K_{x}$ and $K_{y}$ of the examples in the two views $X$ and $Y$ of the data. In contrast to linear CCA, which works by carrying out an eigen-decomposition of the covariance matrix, the eigenvalue problem for KCCA is given by:
\begin{equation}\label{kcca_3}
    \left(
    \begin{array}{cc}
    0 & K_{x}K_{y}\\
    K_{y}K_{x} & 0
    \end{array}
    \right)
    \left(
    \begin{array}{c}
    \alpha\\
    \beta
    \end{array}
    \right)
    =
    \lambda
    \left(
    \begin{array}{cc}
    K_{x}^{2} & 0\\
    0 & K_{y}^{2}
    \end{array}
    \right)
            \left(
    \begin{array}{c}
    \alpha\\
    \beta
    \end{array}
    \right).
\end{equation}
For the case of a linear kernel, KCCA reduces to the standard CCA.

KCCA can isolate feature space directions that correlate between the two views and might be expected to represent common relevant information; therefore, experiments have shown that KCCA could be an effective preprocessing step to improve the performance of classification algorithms such as Support Vector Machine (SVM). Combining KCCA with SVM into a single optimization, \cite{farquhar2005two} proposed a method called SVM-2K, which can be seen as the global optimization of two distinct SVMs, one in each of the two feature spaces. Slightly different from the 2-norm that characterizes KCCA, SVM-2K takes an $\epsilon$-insensitive 1-norm using slack variable to measure the amount by which points fail to meet $\epsilon$ similarity:
\begin{equation}\label{}\nonumber
    |\langle \textbf{w}_{A},\phi_{A}(\textbf{x}_{i})\rangle+b_{A}-\langle \textbf{w}_{B},\phi_{B}(\textbf{x}_{i})\rangle-b_{B}|\leq \eta_{i}+\epsilon{} ,
\end{equation}
where $\textbf{w}_{A}, b_{A}$ and $\textbf{w}_{B}, b_{B}$ are the weight and bias of the first and second SVM respectively. Then with the usual 1-norm SVM constraints, the objective problem can be written as:
\begin{eqnarray}
  \min \quad L &=& \frac{1}{2}\|\textbf{w}_{A}\|^{2}+\|\textbf{w}_{B}\|^{2}\\\nonumber
  &&
  +C^{A}\sum_{i}\xi_{i}^{A}+C^{B}\sum_{i}\xi_{i}^{B}
  +D\sum_{i}\eta_{i} \\\nonumber
  \texttt{s.t.} && |\langle \textbf{w}_{A},\phi_{A}(\textbf{x}_{i})\rangle+b_{A}-\langle \textbf{w}_{B},\phi_{B}(\textbf{x}_{i})\rangle-b_{B}|\leq \eta_{i}+\epsilon \\\nonumber
   && y_{i}(|\langle \textbf{w}_{A},\phi_{A}(\textbf{x}_{i})\rangle +b_{A})\geq 1-\xi_{i}^{A} \\\nonumber
   && y_{i}(|\langle \textbf{w}_{B},\phi_{B}(\textbf{x}_{i})\rangle+b_{B})\geq 1-\xi_{i}^{B} \\\nonumber
   && \xi_{i}^{A}\geq 0,\quad \xi_{i}^{B}\geq 0,\quad \eta_{i}\geq 0 \;\forall i.
\end{eqnarray}
The final decision function is
\begin{equation}\label{}
    f(x)= \frac{1}{2}(f_{A}(x)+f_{B}(x)).
\end{equation}

\subsubsection{Theoretical analysis of CCA}

Canonical correlation analysis (CCA) can be viewed as finding basis vectors for two sets of variables such that the correlations between the projections onto these basis vectors $x_{a}=w_{a}^{T}\phi_{a}(x)$ and $y_{b}=w_{b}^{T}\phi_{b}(y)$ are mutually maximized. KCCA uses the kernel trick to produce a non-linear version of CCA, by looking for functions $f\in \textit{H}_{x}$ and $g\in \textit{H}_{y}$ such that the random variables $f(x)$ and $g(y)$ have maximal correlation. This leads to the kernelised form, KCCA
\begin{equation}\label{kcca_bound}
    \max \frac{Cov[f(x),g(y)]}{Var[f(x)]^{1/2}Var[g(y)]^{1/2}}.
\end{equation}
In practice, we have to estimate the desired function from a finite sample, thus an empirical estimate of Eq. (\ref{kcca_bound}) is
\begin{equation}\label{}
    \max \frac{\widehat{Cov}[f(x),g(y)]}{(\widehat{Var}[f(x)]^{1/2}+\varepsilon_{n}\|f\|^{2}_{H_{x}})(\widehat{Var}[g(y)]^{1/2}+\varepsilon_{n}\|g\|^{2}_{H_{y}})},
\end{equation}
where $\varepsilon_{n}$ is the regularization coefficient and n is the number of examples. \cite{fukumizu2007statistical} investigated the general problem of establishing a consistency of KCCA by providing the rates for the regularization parameter, and  proved that when
\begin{displaymath}
\lim_{n\rightarrow \infty} \epsilon_{n}=0, \; \lim_{n\rightarrow \infty} \frac{n^{-1/3}}{\epsilon_{n}}=0,
\end{displaymath}
for the decay of the regularization coefficient $\varepsilon_{n}$, the convergence in the $L_{2}$ norm for kernel CCA is ensured.

\cite{hardoon2009convergence} proposed a finite sample statistical analysis of KCCA by using a regression formulation. By computing the empirical expected value of $g_{a,b}(x,y):=\widehat{E}[\|W_{a}^{T}\phi_{a}(x)-W_{b}^{T}\phi_{b}(y)\|^{2}]$,, the error bound on new data can be obtained by using Rademacher complexity. Formally, given a paired training set $S=\{(x_{i},y_{i})\}$ of size $\mathcal{L}$ in the feature space defined by the bounded kernels $k_{a}$ and $k_{b}$ drawn i.i.d according to a distribution $\mathcal{D}$, then with probability greater than $1-\delta$ over the generation of $S$, the expected value of $g_{a,b}(x,y)$ on new data is bounded by
\begin{eqnarray}
  E_{D}[g_{a,b}] &\leq& \widehat{E}_{D}[g_{a,b}]+3RA\sqrt{\frac{\ln{2/\delta}}{2\mathcal{L}}}  \\\nonumber
   &+& 4A\frac{1}{\mathcal{L}} \sqrt{\sum_{i=1}(k_{a}(x_{i},x_{i})+k_{b}(y_{i},y_{i}))^{2}},
\end{eqnarray}
where
\begin{displaymath}
R=\max_{x\in \mathcal{D}}(k_{a}(x,x)+k_{b}(y,y))
\end{displaymath}
\begin{displaymath}
\|W_{a}^{T}W_{a}+W_{b}^{T}W_{b}\|^{2}\leq A.
\end{displaymath}
This suggests the regularization of KCCA because it shows that the quality of the generalization of the associated pattern function is controlled by the sum of the squares of the norms of the weight vectors.

\cite{cai2011convergence} gave a convergence rate analysis of kernel CCA. Assuming $(\mathcal{H}_{x}, \mathcal{H}_{y})$ are RKHS of functions on  $\mathcal{X}$ and $\mathcal{Y}$ respectively, $V_{YX}$ is a compact operation from $\mathcal{H}_{x}$ to $\mathcal{H}_{y}$, and there exist operators $W_{l},W_{r}$ such that
\begin{displaymath}
V_{YX}=W_{l}\Sigma^{p}_{XX} \; \texttt{and} \; V_{XY}=\Sigma^{p}_{YY}W_{r},
\end{displaymath}
where $\Sigma^{p}_{XX}$ and $\Sigma^{p}_{XX}$ are covariance operators. Taken $\varepsilon_{n}=\varepsilon_{1}n^{-\alpha}$ with $0<\alpha<1/3$, then with probability at least $1-\delta$, we have
\begin{equation}\label{}
    \|\Sigma^{1/2}_{XX}(\widehat{f}_{n}-\widehat{f})\|_{H_{x}}^{2}\leq C_{6,\delta}n^{-\theta},\;
    \|\Sigma^{1/2}_{YY}(\widehat{g}_{n}-\widehat{g})\|_{H_{y}}^{2}\leq C_{6,\delta}n^{-\theta},\;
\end{equation}
where $\theta=\min\{1-3\alpha,2p\alpha,\alpha\}$ and $C_{6,\delta}$ is a constant independent of $n$. So when $0<p\leq \frac{1}{2}$, the convergence rate is $\min\{1-3\alpha,2p\alpha,\alpha\}$.

\subsubsection{Related algorithms with CCA}

CCA has been widely studied in different fields as a general tool for conducting multi-view dimensional reduction. Recently many new algorithms based on CCA have been proposed to extend the original CCA in different applications.

One popular use of CCA is for supervised learning, in which one view is derived from the data and another view is derived from the class labels. In this setting, the data can be projected into a lower-dimensional space directed by the label information \citep{yu2006multi}. However, this algorithm does not actually use the multiple views of the data; it is just a single view approach along with the label information. \cite{gma} proposed a Generalized Multi-view Analysis (GMA) which exploits the fact that most popular supervised and unsupervised feature extraction techniques are the solution of a special form of quadratically constrained quadratic program. This algorithm can be seen as a supervised extension of CCA and has the potential to replace CCA whenever classification or retrieval is the purpose and label information is available.

\cite{muliview_clustering_cca} exploited CCA to project the data to the subspace spanned by the means, and then applied standard clustering algorithms to this subspace. This subspace is valuable for the subsequent clustering, because, when projected onto this subspace, the means of the distributions are well-separated, yet the typical distance between points from the same distributions is smaller than in the original space. Both traditional CCA and KCCA assume that features across all views are available for examples, but this may not be the case with many multi-view datasets. To apply multi-view clustering on such datasets, \cite{multiview_imcomplete_view} found a way to deal with the lack of data in the incomplete views with an idea from Laplacian regularization. Given the known part of $K$, the missing parts of kernel matrix$K$ can be found by solving an optimization problem; following construction of the full kernel, standard algorithms can conduct the subsequent tasks.

In semi-supervised learning, a number of labeled examples are usually required for training an initial weakly useful predictor which is in turn used to exploit the unlabeled examples. By taking advantage of the correlations between the views using CCA, \cite{few_labeled_examples} proposed a method which can perform semi-supervised learning with only one labeled training example. With the help of CCA, the similarity between an original unlabeled instance and the original labeled instance can be measured. Thus, several unlabeled examples with highest and lowest similarity scores can be selected as the extra positive and negative examples, respectively. As the number of labeled training examples is increased, the traditional semi-supervised learning algorithm can be performed.

\cite{mkl_cca} developed a novel multiple kernel learning algorithm, combined with CCA. Initially the input data is mapped into m different feature spaces by m different kernels, where each generated feature space is taken as one view of the input data. Borrowing the motivating argument from CCA that m views in the transformed coordinates can be maximally correlated, the generalization of classifiers can be improved. Combining CCA with PCA, \cite{dr_cca_pca} suggested a novel method called MKCCA to implement dimensionality reduction. MKCCA improves the kernel CCA by performing PCA followed by CCA to better remove noises and handle the issue of trivial learning. Furthermore, comparing CCA with least squares for regression and classification, \cite{least_squares_cca} formulated CCA in multi-label classification as a least square problem.

\subsection{Multi-view Fisher Discriminant Analysis}

n contrast to CCA, which ignores label information, \cite{fda} generalized Fisher's discriminant analysis to find informative projections for multi-view data in a supervised setting.

\subsubsection{Two view Fisher Discriminant Analysis}

Given examples drawn from two views of the same underlying semantic object, denoted as $X_{a}$ and $X_{b}$ respectively, the two view Fisher discriminant chooses two sets of weights $w_{a}$ and $w_{b}$ to solve the following optimization problem
\begin{equation}\label{2fda}
\nonumber
    \rho= \frac{w_{a}^{T}X_{a}^{T}yy^{T}X_{b}^{T}w_{b}}{\sqrt{(w_{a}^{T}X_{a}^{T}BX_{a}w_{a}+\mu\|w_{a}\|^{2})\cdot
    (w_{b}^{T}X_{b}^{T}BX_{b}w_{b}+\mu\|w_{b}\|^{2})}}{,}
\end{equation}
where $w_{a}$ and $w_{b}$ are the weight vectors for each view. Since the equation is not affected by rescaling of $w_{a}$ or $w_{b}$, the optimization can be subjected to the following constraints
\begin{eqnarray}
  \nonumber
  w_{a}^{T}X_{a}^{T}BX_{a}w_{a}+\mu\|w_{a}\|^{2} &=& 1, \\
  \nonumber
  w_{b}^{T}X_{b}^{T}BX_{b}w_{b}+\mu\|w_{b}\|^{2} &=& 1.
\end{eqnarray}
The corresponding Lagrangian for this optimization can be written as
\begin{eqnarray}
  \nonumber
  L = w_{a}^{T}X_{a}^{T}yy^{T}X_{b}^{T}w_{b}-\frac{\lambda_{a}}{2}(w_{a}^{T}X_{a}^{T}BX_{a}w_{a}+\mu\|w_{a}\|^{2}-1) - \frac{\lambda_{b}}{2}(w_{b}^{T}X_{b}^{T}BX_{b}w_{b}+\mu\|w_{b}\|^{2}-1),
\end{eqnarray}
which can be solved by differentiating with respect to the weight vectors $w_{a}$ and $w_{b}$.

\subsubsection{Kernel two view Fisher Discriminant Analysis}

By introducing two dual weight vectors $w_{a}=X_{a}^{T}\alpha$ and $w_{b}=X_{b}^{T}\beta$, we have
\begin{equation}\label{}
\nonumber
     \rho= \frac{\alpha X_{a}X_{a}^{T}yy^{T}X_{b}^{T}X_{b}^{T}\beta}{\sqrt{(\alpha X_{a}X_{a}^{T}BX_{a}X_{a}^{T}\alpha+\kappa\|w_{a}\|^{2})\cdot
    (\beta X_{b}X_{b}^{T}BX_{b}X_{b}^{T}\beta+\kappa\|w_{b}\|^{2})}},
\end{equation}
and Its kernel form
\begin{equation}\label{}
\nonumber
     \rho= \frac{\alpha K_{a}yy^{T}K_{b}\beta}{\sqrt{(\alpha K_{a}BK_{a}\alpha+\kappa\|w_{a}\|^{2})\cdot
    (\beta K_{b}BK_{b}\beta+\kappa\|w_{b}\|^{2})}}.
\end{equation}
Given the constraints
\begin{eqnarray}
\nonumber
  \alpha K_{a}BK_{a}\alpha + \kappa \alpha K_{a}\alpha&=& 1, \\\nonumber
  \beta K_{b}BK_{b}\beta +\kappa \beta K_{b}\beta&=& 1,
\end{eqnarray}
the corresponding Lagrangian for this optimization can be written as
\begin{eqnarray}
  \nonumber
  L = \alpha K_{a}yy^{T}K_{b}\beta&-&\frac{\lambda_{a}}{2}(\alpha K_{a}BK_{a}\alpha + \kappa \alpha K_{a}\alpha-1)
   - \frac{\lambda_{b}}{2}(\beta K_{b}BK_{b}\beta +\kappa \beta K_{b}\beta-1).
\end{eqnarray}
Differentiating with respect to the weight vectors $\alpha$ and $\beta$, the above problem can then be solved.

\subsection{Multi-view Embedding}

Since high dimensionality, i.e. a large amount of input features, may lead to a large variance of estimates, noise, over-fitting, and in general, higher complexity and inefficiency in the learners, it is necessary to conduct dimensional reduction and generate  low-dimensional representations for these features. When faced with multiple features, however, performing a dimensional reduction for each feature is not an ideal solution, considering the underlying connections between them. Thus it may be necessary to resort to advanced methods to conduct embedding for multiple features simultaneously and to output a meaningful low-dimensional embedding shared by all features.

Existing spectral embedding algorithms assume that samples are drawn from a vector space and thus cannot deal straightforwardly with multi-view data. \cite{xia2010multiview} developed a new spectral embedding algorithm, namely, multi-view spectral embedding (MSE), which encodes multi-view features to achieve a physically meaningful embedding. Based on their previous work of patch alignment \citep{zhang2009patch}, MSE can be described as follows. MSE first builds a patch for a sample on a view, then given the patches from different views, part optimization is performed to obtain the optimal low-dimensional embedding for each view. All low-dimensional embeddings from different patches are then unified into one whole by global coordinate alignment. More formally, given the $i$-th view $X^{i}=[x^{i}_{1},\cdots,x^{i}_{n}]$, consider an arbitrary point $x_{j}^{i}$ and its $k$ nearest neighbors, $x_{j}^{i}$ is defined as $X_{j}^{i}=[x_{j}^{i},x_{j1}^{i},\cdots,x_{jk}^{i}]$. For $X_{j}^{i}$, we want to find a part mapping $f_{j}^{i}:X_{j}^{i}\rightarrow Y_{j}^{i}$, where $Y_{j}^{i}=[y_{j}^{i},y_{j1}^{i},\cdots,y_{jk}^{i}]$. The part optimization for the $j$-th patch on the $i$-th view is defined as
\begin{equation}\label{mse1}
    \min_{Y_{j}^{i}} \quad \sum_{l=1}^{k}\|y_{j}^{i}-y_{jl}^{i}\|^{2}(w_{j}^{i})_{l},
\end{equation}
where $(w_{j}^{i})_{l}=exp(-\|x_{j}^{i}-x_{jl}^{i}\|^{2}/t)$. Eq. (\ref{mse1}) can be reformulated to
\begin{equation}\label{}
     \min_{Y_{j}^{i}} \quad \texttt{tr}(Y_{j}^{i}L_{j}^{i}(Y_{j}^{i})^{T}),
\end{equation}
where $\texttt{tr}(\cdot)$ is the trace operator and $L_{j}^{i}$ encodes the objective function for the $j$-th patch on the $i$-th view. To explore the complementary property of multiple views, a set of non-negative weights $\alpha = [\alpha_{1},\cdots,\alpha_{m}]$ is imposed on part optimizations, thus the multi-view part optimization for the $j$-th patch is
\begin{equation}\label{}
     \min_{\{Y_{j}^{i}\}_{i=1}^{m}} \quad \sum_{i=1}^{m}\alpha_{i}\texttt{tr}(Y_{j}^{i}L_{j}^{i}(Y_{j}^{i})^{T}).
\end{equation}
To ensure that low dimensional embeddings in different views are globally consistent with each one another, assume that the coordinate for $Y_{j}^{i}=[y_{j}^{i},y_{j1}^{i},\cdots,y_{jk}^{i}]$ is selected from the global coordinate $Y=[y_{1},\cdots,y_{n}]$, which then gives $Y_{j}^{i}=YS_{j}^{i}$, where $S_{j}^{i}$ is the selection matrix for encoding the relationships of samples in a patch in the original high dimensional space. By summing over all part optimizations, the global coordinate alignment can be written as
\begin{equation}\label{mse8}
    \min \quad \sum_{j=1}^{n}\sum_{i=1}^{m}\alpha_{i}\texttt{tr}(YS_{j}^{i}L_{j}^{i}(S_{j}^{i})^{T}Y^{T}).
\end{equation}
From Eq. (mse8), the alignment matrix for the i-th view can be written as
\begin{equation}\label{}
    L^{i}=\sum_{j=1}^{n}S_{j}^{i}L_{j}^{i}(S_{j}^{i})^{T}.
\end{equation}
To make sure that each view makes a particular contribution to the final low dimensional embedding, and considering some constraints on the variants, the final objective function is defined as
\begin{eqnarray}
  \min_{Y,\alpha} && \sum_{i=1}^{m}\alpha_{i}^{\gamma}\texttt{tr}(YL^{i}Y_{T}) \\\nonumber
  \texttt{s.t.} && YY^{T}=I,\; \sum_{i=1}^{m}\alpha_{i}=1,\; \alpha_{i}\geq 0, \;  \gamma >1.
\end{eqnarray}
Finally, MSE can generate a low dimensional sufficiently smooth embedding by preserving the locality of each view simultaneously.

The main idea of Stochastic Neighbor Embedding (SNE) is to construct probability distributions from pair wise distances wherein larger distances correspond to smaller probabilities and vice versa. Formally, suppose we have high-dimensional data points $\{x_{i}\}_{i=1}^{n}$, the joint probability distribution over sample pairs can be represented in a symmetric matrix $P\in \mathbb{R}^{n\times n}$, where $p_{ii}=0$ and $\sum_{i,j}p_{ij}=1$. Let $y_{i}$ be the low dimensional data corresponding to $x_{i}$, then the probability distribution $Q$ in low dimensional embedding is defined as
\begin{equation}\label{}
    q_{ij}=\frac{(1+\|y_{i}-y_{j}\|^{2})^{-1}}{\sum_{k\neq l}(1+\|y_{k}-y_{l}\|^{2})^{-1}}.
\end{equation}
This embedding can be acquired by minimizing the KL divergence of the two probability distributions,
\begin{equation}\label{}
    \texttt{KL}(P|Q) = \sum_{i\neq j}p_{ij}\log\frac{p_{ij}}{q_{ij}}.
\end{equation}
\cite{xie2011m} proposed the m-SNE algorithm to generalize SNE to handle multi-view data by introducing one combination coefficient to each view. The final probability distribution on the high dimensional space is then
\begin{equation}\label{}
    p_{ij}=\sum_{t=1}^{v}\alpha^{t}p_{ij}^{t},
\end{equation}
where $\alpha^{t}$ is the combination coefficient for view $t$ and $p_{ij}^{t}$ is the probability distribution on view $t$. This combination coefficient plays an important role in utilizing the complementary information and suppressing noise in multi-view data. Additionally, the original objective function contains only KL divergence; a 2-norm regularization term is added to balance the coefficients over all views
\begin{equation}\label{}
    g(\alpha) = \sum_{i\neq j}p_{ij}\log\frac{p_{ij}}{q_{ij}} +\lambda \|\alpha\|^{2},
\end{equation}
where $\lambda$ is the tradeoff coefficient.

\cite{hansparse} proposed a new framework of sparse unsupervised dimensionality reduction for multi-view data. Considering the specific statistical property of each view, this algorithm first learns low-dimensional patterns from these views using the principal component analysis (PCA) algorithm. After combining the learned low-dimensional pattern of each view into one unified pattern, the construction of the low-dimensional consensus representation can be formulated to approximate the matrix of patterns by means of a low-dimensional consensus base matrix and a loading matrix. To select the most discriminative feature for the spectral embedding of multiple views, a 1-norm is added into the loading matrix's columns and orthogonal constraints are imposed on the base matrix. A novel method called Spectral Sparse Multi-View Embedding (SSMVE) was subsequently developed to efficiently obtain the solution. Furthermore, since each row of the loading matrix is a vector concatenated by several parts which correspond to the different patterns learned from different views, a novel structured sparsity-inducing norm penalty was imposed on the loading matrix's rows to gain flexibility in sharing information across subsets of the views. Consequently, another approach for multi-view dimensionality reduction with structured sparsity penalty, namely, Structured Sparse Multi-View Dimensionality reduction (SSMVD), was proposed.

\subsection{Multi-view Metric Learning}

The goal of metric learning for multi-view data is to construct embedding projections from the data in different representations into a shared feature space, so that the Euclidean distance in this space is meaningful not only within a single view, but also between different views.

Motivated by cross-media retrieval tasks, \cite{quadrianto2011learning} studied the metric learning problem to find the joint Euclidean distance function to allow nearest neighbor queries. Following the classical principle of pulling samples together if they are related and pushing them apart if they are not, multi-view metric learning is formulated as follows. Suppose there are two sets of $m$ data points, $X=\{x_{1},\cdots,x_{m}\}$ and $Y=\{y_{1},\cdots,y_{m}\}$ describing the same objects from two different views, and for each $x_{i}\in X$ there exists a set $S_{x_{i}}$ of data points from $Y$ which are similar to $x_{i}$. Given $X=\mathbb{R}^{d_{1}}$ and $Y=\mathbb{R}^{d_{2}}$, we seek the projection functions,
\begin{equation}\label{}\nonumber
    g_{1}:\mathbb{R}^{d_{1}}\longrightarrow \mathbb{R}^{D} \quad and \quad g_{2}:\mathbb{R}^{d_{2}}\longrightarrow \mathbb{R}^{D},
\end{equation}
with $D\ll min(d_{1},d_{2})$ that respects the neighborhood relationship $\{S_{x_{i}}\}_{i=1}^{m}$. Considering a linear parameterization of the functions $g_{1}(x_{i})=\langle w_{1},\phi(x_{i})\rangle$ and $g_{2}(y_{i})=\langle w_{2},\phi(y_{i})\rangle$, then the metrics $w_{1}$ and $w_{2}$ are the goal of the learning, and the objective function can be written as
\begin{eqnarray}
  L(w_{1},w_{2},X,Y,S) = \sum_{i,j=1}^{m}L^{i,j}(w_{1},w_{2},x_{i},y_{j},S_{x_{i}})
   + \eta \Omega (w_{1})+\gamma \Omega (w_{2}),
\end{eqnarray}
where $L^{i,j}(\cdot)$ is the loss function, $\Omega(\cdot)$ is a regularizer on the parameters and $\eta$ and $\gamma$ are trade-off variables. By choosing the loss function appropriately, the properties the projected data are expected to have can be expressed. In particular, if it is hoped to ensure that similar objects across different views are mapped to nearby points, whereas dissimilar objects across different views are to be pushed apart, the loss function can be designed as the union of two different parts,
\begin{equation}\label{}
    L(w_{1},w_{2},X,Y,S)=\frac{\textbf{I}_{y_{i}\in S_{x_{i}}}}{2}\times L_{1}^{i,j}+\frac{1-\textbf{I}_{y_{i}\in S_{x_{i}}}}{2}\times L_{2}^{i,j},
\end{equation}
where the similarity term $L_{1}^{i,j}$ forces similar objects to be at proximal locations in the latent space and the dissimilar term $L_{2}^{i,j}$ pushes dissimilar objects away from one another. This objective function can be decomposed into a difference of two concave functions, thus it can be solved efficiently by the concave convex procedure (CCCP).

Since various different low-level visual features can be extracted to comprehensively represent the image in image processing, it is difficult to choose which feature to depend on to measure the similarity between images. Thus \cite{yu2012semi} proposed a semi-supervised multi-view distance metric learning (SSM-DML) algorithm to construct an accurate metric to precisely measure the dissimilarity between different examples associated with multiple views. Formally define a matrix $\textbf{F}=[\textbf{F}_{1}^{T},\cdots,\textbf{F}_{N}^{T}]^{T}$, where $F_{ij}$ is the confidence of $x_{i}$ with the label $y_{j}$, and then this matrix $\textbf{F}$ can be obtained by minimizing the following objective function:
\begin{equation}\label{llgc}
    Q = \sum_{i,j=1}^{N}\textbf{W}_{ij}\|\frac{\textbf{F}_{i}}{\sqrt{\textbf{D}_{ii}}}-\frac{\textbf{F}_{j}}{\sqrt{\textbf{D}_{jj}}}\|^{2}+\mu \sum_{i=1}^{N}\|\textbf{F}_{i}-\textbf{Y}_{i}\|^{2},
\end{equation}
where $\textbf{W}$ is an affinity matrix with $W_{ij}$ indicating the dissimilarity measure between $x_{i}$ and $x_{j}$, and $\textbf{D}$ is a diagonal matrix with $D_{ii}$ equal to the sum of the $i$-th row of $\textbf{W}$. The first term in Eq. (\ref{llgc}) implies the smoothness of the labels on the graph and the second term indicates the constraint of the training data. Suppose $X^i$ represents the $i$-th view of the example; by linearly combining the graphs constructed from multi-view features sets through the weights $\alpha$, Eq. (\ref{llgc}) can be extended to the multi-view feature sets
\begin{eqnarray}
  Q &=& \sum_{k=1}^{K}\sum_{i,j=1}^{N}\alpha_{k}\textbf{W}_{ij}^{k}\|\frac{\textbf{F}_{i}}{\sqrt{\textbf{D}_{ii}^{k}}}-\frac{\textbf{F}_{j}}{\sqrt{\textbf{D}_{jj}^{k}}}\|^{2}+\mu \sum_{i=1}^{N}\|\textbf{F}_{i}-\textbf{Y}_{i}\|^{2}+\lambda \|\alpha\|^{2}\\\nonumber
  \texttt{s.t.} && \sum_{k=1}^{K}\alpha_{k}=1.
\end{eqnarray}
Then through adopting alternating optimization to solve the above optimization problem, SSM-DML can learn the multi-view distance metrics from multiple feature sets and the labels of unlabeled data simultaneously.

\cite{zhai2012multiview} also studied the multi-view metric learning problem in the semi-supervised learning setting, and proposed a new method called Multi-view Metric Learning with Global consistency and Local smoothness (MVML-GL), which jointly considers global consistency and local smoothness. This algorithm is accomplished in two steps: (1) seek a shared latent feature space to establish the relationship between data from multi-view observation spaces according to pairs of labeled instances; (2) learn the relationships between the input space of each observation and the shared latent space for unlabeled and test data. It is worth noting that this first step is globally consistent, as it simultaneously considers the geometric structures contained in each view and connections between the data from different views, and the second step is locally smooth, which enables each instance to have its own specific distance metric instead of applying a uniform metric for all instances. Additionally, both steps can be formulated as convex optimization problems with closed form solutions, thus they can be efficiently solved.

\subsection{Latent Space Models}

Besides the aforementioned methods, which aim to conduct meaningful dimensional reduction for multi-view data, there are also works that concentrate on analyzing the relationships between different views. These methods are used to build latent space models, with which multiple views can be connected with one another through latent variables, and the information can be propagated from one view to another view.

\subsubsection{Shared Gaussian Process Latent Variable Model}

Gaussian processes (GPs) are powerful models for classification and regression that subsume numerous classes of function approximators, such as single hidden-layer neural networks and RBF networks. \cite{gplvm} first proposed the Gaussian process latent variable model (GPLVM) as a new technique for non-linear dimensional reduction. \cite{sgplvm} proposed the shared GPLVM (SGPLVM) as a generalization of the GPLVM model that can handle multiple observation spaces, where each set of observations is parameterized by a different set of kernel parameters.

Let $Y, Z$ be matrices of observations drawn from spaces of dimensionality $D_{Y}, D_{Z}$ respectively, and $X$ be a latent space of dimensionality $D_{X}\ll D_{Y}, D_{Z}$. Assume that each latent point $x_{i}$ generates a pair of observations $y_{i},z_{i}$ via GPs parameterized non-linear functions $f_{Y}:X\rightarrow Y$ and $f_{Z}:X\rightarrow Z$. By using an exponential (RBF) kernel to define the similarity between two data points $x, x^{'}$
\begin{equation}\label{}
    k(x,x^{'})=\alpha_{Y}exp(-\frac{\gamma_{Y}}{2}\|x-x^{'}\|^{2})+\delta_{x,x^{'}}\beta_{Y}^{-1},
\end{equation}
the priors $P(\theta_{Y}),P(\theta_{Z}),P(\theta_{X})$ ($\theta=\{\alpha,\beta,\gamma\}$) and the likelihoods $P(Y),P(Z)$ for the $Y,Z$ observation spaces are given by
\begin{equation}\label{}
    P(Y|\theta_{Y},X)=\frac{|W|^{N}}{\sqrt{(2\pi)^{ND_{Y}}|K|^{D_{Y}}}}exp(-\frac{1}{2}\sum_{k=1}^{D_{Y}}w_{k}^{2}Y_{k}^{T}K_{Y}^{-1}Y_{k}),
\end{equation}
\begin{equation}\label{}
    P(Z|\theta_{Z},X)=\frac{|W|^{N}}{\sqrt{(2\pi)^{ND_{Z}}|K|^{D_{Z}}}}exp(-\frac{1}{2}\sum_{k=1}^{D_{Z}}w_{k}^{2}Z_{k}^{T}K_{Z}^{-1}Z_{k}),
\end{equation}
\begin{equation}\label{}
    P(\theta_{Y})\propto \frac{1}{\alpha_{Y}\beta_{Y}\gamma_{Y}} \quad P(\theta_{Z})\propto \frac{1}{\alpha_{Z}\beta_{Z}\gamma_{Z}},
\end{equation}
\begin{equation}\label{}
    P(X)=\frac{1}{\sqrt{2\pi}}exp(-\frac{1}{2}\sum_{i}\|x_{i}\|^{2}),
\end{equation}
then the joint likelihood can be written as
\begin{equation}\label{eq:sgplvm5}
P_{GP}(X,Y,Z,\theta_{Y},\theta_{Z})=P(Y|\theta_{Y},X)P(Z|\theta_{Y},X)P(\theta_{y})P(\theta_{Z})P(X).
\end{equation}
By using a conjugate gradient solver to maximize Eq. (\ref{eq:sgplvm5}), the model can learn a separate kernel for each observation space and a single set of common latent points.

Given a trained SGPLVM, we would like to infer the parameters in one observation space given the parameters in the other observation space. This problem can be solved in two steps. First, we determine the most likely latent coordinate $x$ given the observation $y$ using $\max_{x}L_{x}(x,y)$. Once the correct latent coordinate $x$ has been inferred for a given $y$, the model uses the trained SGPLVM to predict the corresponding observation $z$.

\subsubsection{Shared Kernel Information Embedding}

Given samples drawn from a distribution p(x), Kernel Information Embedding \citep{kie} aims to find a low-dimensional latent distribution, $p(z)$, that captures the structure of the data, along with explicit bidirectional probabilistic mappings between the latent space and the data space. In particular, KIE finds the joint distribution $p(x,z)$ that maximizes the mutual information between the latent distribution and the data distribution:
\begin{eqnarray}
    I(x,z)&=&\int p(x,z)log\frac{p(x,z)}{p(x)p(z)}dxdz\\
          \nonumber
          &=& H(x)+H(z)-H(x,z),
\end{eqnarray}
where $H(\cdot)$ is the usual Shannon entropy, which can be estimated by kernel density.

The shared KIE (sKIE)\citep{skie, memisevic2012shared}, which can be seen as the extension of KIE, constructs the joint embedding for two views by maximizing the mutual information $I((x,y),z)$. Assuming the conditional independence of $x$ and $y$ given $z$, $I((x,y),z)$ can be expressed as a sum of two mutual information terms,
\begin{equation}
    I((x,y),z)=I(x,z)+I(y,z),
\end{equation}
where $I(x,z)$ and $I(y,z)$ can be formulated as KIE.

An application of this algorithm is human pose inference. For discriminative pose inference, the aim is to find likely poses $y$ conditioned on input image features $x^{*}$. Then the conditional pose distribution is:
\begin{equation}
    p(y|x^{*})=\int_{z}p(y|z)p(z|x^{*})dz.
\end{equation}
Alternatively, the focus can be on identifying the principal modes of $p(y|x^{*})$. To this end, it is assumed that the principal modes of $p(y|x^{*})$ coincide with the principal modes of the conditional latent distribution $p(z|x^{*})$. That is, a search is first conducted for local maxima of $p(z|x^{*})$, denoted $\{z_{k}^{*}\}_{k=1}^{K}$ for $K$ modes. From these latent points it is straightforward to perform either MAP inference or take the expectation over the conditional pose distributions $p(y|z_{k}^{*})$.

\subsubsection{Factorized Orthogonal Latent Space}

Both sGPLVM and sKIE only consider the shared information in the views of data but ignore the private part in each view. \cite{fols} proposed a robust approach called FOLS to factorize the latent space into shared and private spaces by introducing orthogonality constraints, which penalize redundant latent representations.

For minimal factorization, the shared and private latent spaces are required to be non-redundant; in other words, it is desirable to penalize the redundancy of different private spaces and thus encourage the representation of common information in the shared space. More formally, define $Y^{i}=[y_{1}^{i},\cdots,y_{N}^{i}]^{T}$ as the set of observations from a single view $i$, with $1\leq i \leq V$. Additionally, let $X=[x_{1},\cdots,x_{N}]^{T}$ be the latent space shared across different views, $Z^{i}=[Z_{1}^{i},\cdots,Z_{N}^{i}]^{T}$ be the private space for $i$-th view, and $M^{i}=[m_{1}^{i},\cdots,m_{N}^{i}]^{T}$ be the joint shared-private latent space for each view, with $m_{j}^{i}=[x_{j},z_{j}^{i}]$. By imposing the above mentioned non-redundant constraint as a soft penalty, a FOLS model can be learned by minimizing
\begin{eqnarray}
  \mathcal{L} &=& L+\alpha\underbrace{\sum_{i}(\|X^{T}\cdot Z^{i}\|^{2}_{F}+\sum_{j>i}\|(Z^{i})^{T}\cdot Z^{j}\|^{2}_{F})}_{orthogonality} \\
  \nonumber
   &&+\gamma \underbrace{\sum_{i}\phi(s_{i})}_{low\;  dimensionality}+\eta\underbrace{\sum_{i}(E_{0}^{i}-\sum_{j}s_{i,j}^{2})^{2}}_{energy\; conservation},
\end{eqnarray}
where $s_{i}$ are the singular values of $M^{i}$, $E_{0}^{i}$ is the energy of stream $i$, and $L$ is the loss function of the particular model into which the factorization constraints are introduced. In the sGPLVM and sKIE models, $L$ represents the square loss, or the negative mutual information between each joint latent space and its corresponding data stream.

\subsubsection{Factorized Latent Spaces with Structured Sparsity}

Inspired by sparse coding techniques, \cite{fls_sparsity} proposed a novel approach to finding a latent space in which the information is correctly factorized into shared and private parts, while avoiding the computational burden of previous techniques. In particular, this algorithm represents each view as a linear combination of view-dependent dictionary entries. While the dictionaries are specific to each view, the weights of these dictionaries act as latent variables and are the same for all the views.

More formally, to find a shared-private factorization of the latent embedding $\alpha$ that represents the multiple input modalities, the algorithm adopts the idea of structured sparsity and aims to find a set of dictionaries $\mathcal{D}=\{D^{1},\cdots,D^{V}\}$. This problem can be formulated as,
\begin{equation}
    \min_{\mathcal{D},\alpha} \frac{1}{N}\sum_{v=1}^{V}\|X^{v}-D^{v}\alpha\|^{2}_{F}+\lambda\sum_{v=1}^{V}\psi((D^{v})^{T})+\gamma \psi(\alpha),
\end{equation}
where the first item measures the loss, the second item encourages each view to only use a limited number of latent dimensions, and the third item indicates a relaxation of rank constraints to discover the dimensionality of the latent space.

At inference, given a new observation $\{x_{*}^{1},\cdots,x_{*}^{V}\}$, the corresponding latent embedding $\alpha_{*}$ can be obtained by solving the convex problem
\begin{equation}\label{}
    \min_{\alpha_{*}}\sum_{v=1}^{V}\|x_{*}^{v}-D^{v}\alpha_{*}\|^{2}_{2}+\gamma \|\alpha_{*}\|_{1},
\end{equation}
where the regularizer allows us to deal with noise in the observations.

\subsubsection{Latent Space Markov network}

\cite{multi_view_mn} constructed a predictive subspace shared by multi-view data based on the generic multi-view latent space Markov network (MN), under the assumption that the data from different views and the response variables are conditionally independent given a set of latent variables.

The two-view latent space Markov networks consist of two views of input data $\mathbf{X}:\{X_{n}\}$ and $\mathbf{Z}:\{Z_{m}\}$ and a set of latent variables $\mathbf{H}:\{H_{k}\}$. According to random field theory, the marginal distributions for two views respectively can be written in the exponential forms
\begin{eqnarray}
  p(x) &=& \exp\{\sum_{i}\theta_{i}^{T}\phi(x_{i},x_{i+1})-A(\theta)\}, \\
  p(z) &=& \exp\{\sum_{j}\eta_{i}^{T}\psi(z_{j},z_{j+1})-B(\eta)\},
\end{eqnarray}
where $\phi$ and $\psi$ are feature functions, $A$ and $B$ are log partition functions. For the latent variables, the marginal distribution is
\begin{equation}\label{}
    p(h)=\prod_{k}\exp\{\lambda_{k}^{T}\varphi(h_{k})-C_{k}(\lambda_{k})\},
\end{equation}
where $\varphi(h_{k})$ is the feature vector of $h_{k}$, $C_{k}$is the log-partition function. By combining the above components in the log-domain, the joint model distribution is defined as
\begin{equation}\label{}\nonumber
\begin{split}
  p(x,z,h)&  \varpropto\exp\{  \sum_{i}\theta_{i}^{T}\phi(x_{i},x_{i+1})+ \sum_{j}\eta_{i}^{T}\psi(z_{j},z_{j+1})+\lambda_{k}^{T}\varphi(h_{k})\\
    &+ \sum_{ik}\phi(x_{i},x_{i+1})^{T}W_{i}^{k}\varphi(h_{k})
   +\sum_{jk}\psi(z_{j},z_{j+1})^{T}U_{j}^{k}\varphi(h_{k}).
\end{split}
\end{equation}

Additionally considering each input sample is associated with a supervised response variable $y\in\{1,\cdots,T\}$, we can define
\begin{equation}\label{}
    p(y|h)=\frac{exp\{V^{T}f(h,y)\}}{\sum_{y^{'}exp\{V^{T}f(h,y^{'})\}}},
\end{equation}
where $f(h,y)$ is the feature vector whose elements from $(y-1)K+1$ to $yK$ are those of $h$ and all others are 0. Accordingly, $V$ is a stacking parameter vector of $T$ sub-vectors $V_{y}$, each of which corresponds to a class label $y$.

Although this multi-view latent space MNs can be learned by maximum likelihood estimation (MLE), \cite{multi_view_mn} estimated the decision boundary directly in a large margin approach. Assume the discriminant function $F(y,h;V)$ is linear, that is, $F(y,h;V)=V^{T}f(h,y)$, which looks like the discriminant function $W^{T}X$ in SVM. Then the objective function is
\begin{equation}\label{largin_margin}
    \min_{\Theta,V}\quad L(\Theta)+\frac{1}{2}C_{1}\|V\|^{2}+C_{2}\mathcal{R}_{hinge}(V),
\end{equation}
where the first item $L(\Theta)=-\sum_{d}\log{p(x_{d},z_{d})}$ is the negative data likelihood, the second item is the constraint of the decision boundary, and the third item hinge loss acts as the slack variable $\xi$ in SVM. Since Eq. (\ref{largin_margin}) maximizes the data likelihood and minimizes training loss, it can be expected that by solving this problem we can find a predictive latent space representation $p(h|x,z)$ and a prediction model parameter $V$ at the same time.

\section{Applications}
\label{sec:8}

In general, by exploiting the consistency and complement of multiple views, learning models from multi-view data will lead to an improvement in learning performance. Thus multi-view learning has been applied successfully in a number of real-world applications.

Since \cite{Co-training} first proposed the co-training algorithm and applied it to the web document classification problem, this novel method has caught the attention of many researchers and has been widely applied in the field of natural language processing \citep{craven2000learning, muller2002applying, phillips2002exploiting}. \cite{pierce2001limitations} studied the learning behavior of co-training and showed that given a small set of labeled training data and a large set of unlabeled data, co-training can reduce the difference in error between co-trained classifiers and fully supervised classifiers trained on a labeled version of all available data by $36\%$. Unlike previous efforts which cope with the task of word sense disambiguation in a supervised way, \cite{mihalcea2004co} suggested combining co-training with majority voting, with the effect of smoothing the learning curves to improve average performance. \cite{maeireizo2004co} investigated the applicability of co-training to train classifiers that predict emotions in spoken dialogues on features pre-processed in a wrapper approach with forward selection. \cite{kiritchenko2001email, kockelkorn2003using} and \cite{scheffer2004email} treated the email classification problem in the framework of semi-supervised learning, so that the cost of labeling unlabeled data could be eliminated, and a co-training method employed to significantly improve learning performance. Besides these applications involving text or natural language processing, co-training has also found application in the field of computer vision. For instance, \cite{liu2011boosted} studied the human action recognition problem and introduced two new confidence measures, i.e. inter-view confidence and intra-view confidence, to address view sufficiency and view dependency issues in co-training. \cite{christoudias2009co} designed a probabilistic heteroscedastic approach to co-training, which discovers the amount of noise while solving multi-view object recognition tasks. \cite{feng2003bootstrapping} and \cite{feng2004bootstrapping} addressed the image annotating problem by combining co-training with active learning. Thus the requisition for the large labeled training corpus for effective learning is relaxed in co-training and the best examples are selected to label at each stage to maximize the learning objective in active learning. Considering various kinds of visual features, such as color and texture features, as sufficient and uncorrelated views of an image, \cite{zhou2004exploiting} and \cite{cheng2007active} introduced a co-training algorithm to conduct relevance feedback in content-based image retrieval.

As for multiple kernel learning, \cite{kumar2007support, lin2007local} and \cite{varma2007learning} applied it to object classification by linearly combining similarity functions between images so that the combined similarity function yields improved classification performance. \cite{longworth2008multiple} employed multiple kernel learning for object detection with the goal of finding an optimal combination of exponential $\chi^{2}$ kernels, each of which would capture a different feature channel, such as the distribution of edges, and visual words. \cite{kembhavi2009incremental} proposed an incremental multiple kernel learning approach for object recognition. In this case, ``incremental'' means that the images of objects in poses more commonly observed in the scene as well as the kernel weights will be updated in each iteration, thus further improving  the learning performance.

\begin{figure*}[thb]
\begin{center}
\includegraphics[width=\textwidth]{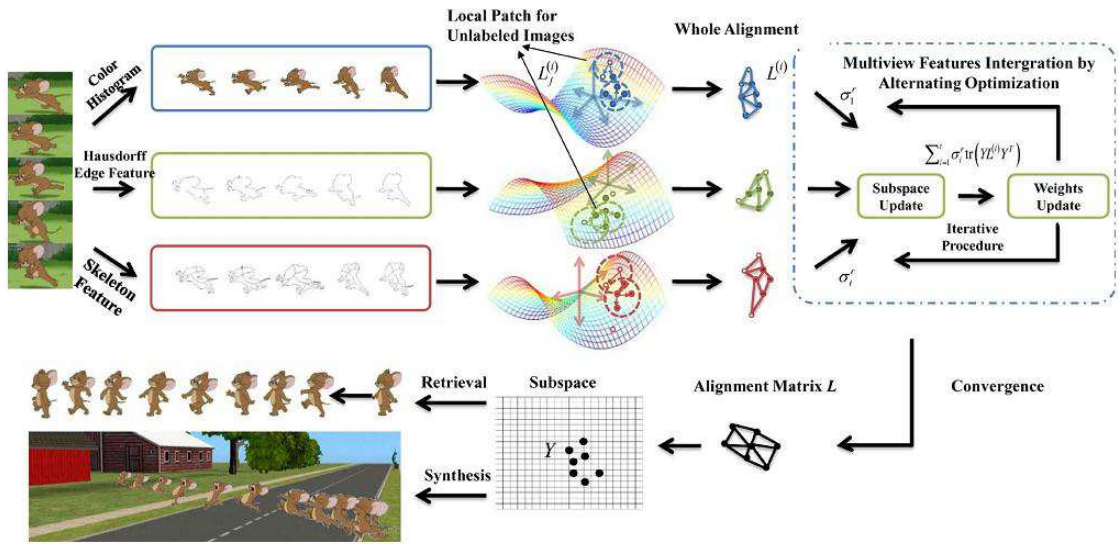}
   \caption{Flowchart of the semi-supervised multi-view subspace learning algorithm \citep{yu2012combining}. The method first extracts multi-view features from cartoon characters. Then, by considering the constraints of each local patch and the complementary characteristics of multi-view features, the low dimensional representation $Y$ can be obtained through solving an alternating optimization problem. Finally, the cartoon character retrieval and clip synthesis can be conducted by measuring the dissimilarity in the subspace $Y$.}\vspace{-5mm}
   \label{fig:5}
\end{center}
\end{figure*}

\begin{figure}[thb]
\begin{center}
\includegraphics[width=\textwidth]{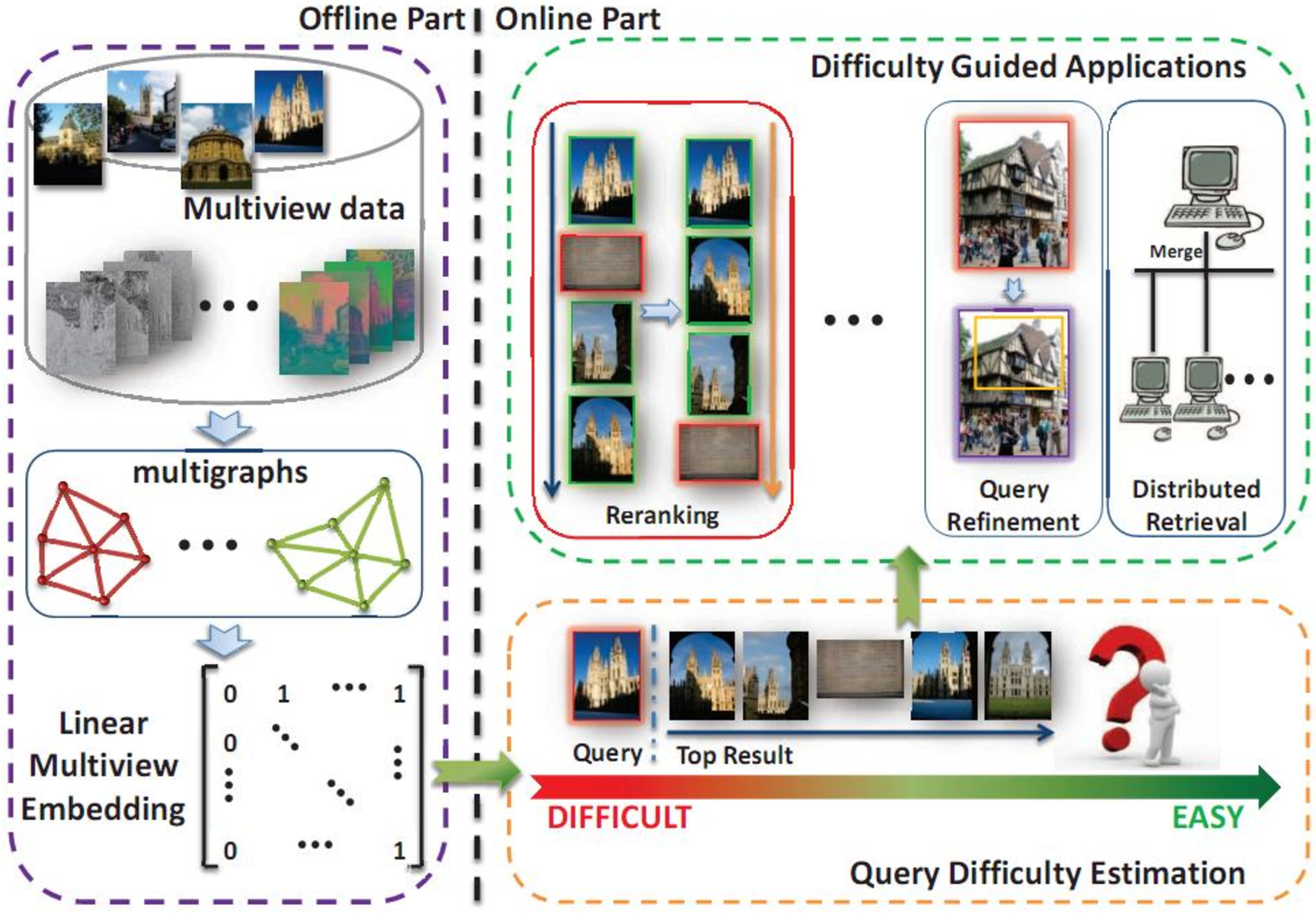}
   \caption{Application of linear multi-view embedding in difficulty-guided image retrieval \citep{li2011difficulty}.}\vspace{-5mm}
   \label{fig:6}
\end{center}
\end{figure}

Subspace learning is an important tool for analyzing the relationships between different views of the data and has a number of applications. \cite{donner2006fast} introduced a fast active appearance model search algorithm based on CCA. \cite{zheng2006facial} used KCCA to solve the facial expression recognition problem. \cite{dhillon2011multi} computed the CCA between different views of the data to estimate low dimensional context specific word representations from unlabeled data in NLP tasks. \cite{fu2008multiple} effectively solved the face recognition task by constructing a linear subspace in which the cumulative canonical correlation between any pair of feature sets is maximized. \cite{zhangcombining} studied the hyperspectral remote sensing image classification problem in the approach of multi-view learning, and introduced the patch alignment framework to linearly combine multiple features in an optimal way and a unified low-dimensional representation of these multiple features for subsequent classification. Considering that the key issue in cartoon character retrieval is proper representation that describes the cartoon character effectively, \cite{yu2012combining} introduced a semi-supervised multi-view subspace learning algorithm which encodes different features in a unified space, as illustrated in Figure \ref{fig:5}. In this unified subspace, the Euclidean distance can be straightforwardly used to measure the distance between two cartoon characters. To improve the performance of the ranking and difficulty estimation in image retrieval, \cite{li2011difficulty} applied multi-view embedding (ME) to images represented by multiple features for integrating a joint subspace by preserving the neighborhood information in each feature space, as illustrated in Figure \ref{fig:6}. To eliminate the ``out of sample'' and huge computation cost problem, a linear multi-view embedding algorithm was developed which learns a linear transformation from a small set of data and can effectively infer the subspace features of new data.

\section{Performance Evaluation}
\label{sec:9}

In this section, we introduce some widely used datasets in multi-view learning experiments and make an empirical comparison of several representative multi-view learning algorithms with single-view learning algorithms.

\textbf{Data Sets for Multi-view Learning}.
So far, several datasets have been widely employed in multi-view learning experiments. Here we give a simple introduction to these datasets.
\begin{itemize}
\item WebKB dataset \footnote{http://www.cs.cmu.edu/afs/cs.cmu.edu/project/\mbox{theo-20}\\/www/data/}  is the most famous dataset used in multi-view learning, on which the co-training algorithm was first evaluated. This dataset consists of 8282 academic web pages collected from computer science department web sites at four universities: Cornell, University of Washington, University of Wisconsin, and University of Texas. These pages can be grouped into six classes: student, staff, faculty, department, course and project. There are two views containing the text on the page and the anchor text of hyperlink respectively.
\item Citeseer dataset \footnote{http://komarix.org/ac/ds/}
    is a collection of scientific publications which contains 3312 documents belonging to six classes. There are three natural views for each document: the text view consists of the title and abstract of the paper; the two link views are inbound and outbound references.
\item Some popular data sets coming from UCI repository \footnote{http://archive.ics.uci.edu/ml/}
    are suitable for evaluating multi-view learning. For example, the internet advertisement dataset contains images from various web pages that are characterized either as advertisements or non-advertisements. The instances are described in terms of six views, which are the geometry of the images, the base url, the image url, the target url, the anchor text and the alt text.
\item There are also a number of other multimedia datasets usually employed in experiments on image annotation, image classification and image retrieval, which include TRECVID2003 video dataset \footnote{http://www-nlpir.nist.gov/projects/tv2003/} , Caltech256 \footnote{http://www.vision.caltech.edu/Image\_Datasets/Caltech256/} , etc. We extract different visual features to represent multiple views of the data, such as color histogram, edge direction histogram, and wavelet texture.
\end{itemize}

\textbf{Empirical Evaluation}.
To illustrate the benefits of multi-view learning methods compared to traditional single-view learning, Table \ref{tab:result_co-training} presents a list drawn from several published multi-view learning papers. \cite{Co-training, Co-em, co_em_svm, Co-regularization, yu2011bayesian} and \cite{dr_cca_pca} used the WebKB data as one of the evaluation datasets. Due to the different preprocessing steps of the algorithms by different researchers, it is difficult to make a direct comparison of the proposed methods; thus we denote them as $\text{WebKB}_{1}, \cdots, \text{WebKB}_{6}$ respectively and show the comparison results between the proposed multi-view learning methods and single-view learning methods in the table.

On the $\text{WebKB}_{1}$ data, \cite{Co-training} evaluated the co-training algorithm and compared its performance with that of the single-view learning algorithm naive Bayes. On the $\text{WebKB}_{2}$,  \cite{Co-em} evaluated the proposed co-EM method. \cite{co_em_svm} developed a novel co-EM based on SVM and showed its satisfactory performance compared to single-view SVM and co-trained naive Bayes on the $\text{WebKB}_{3}$.  \cite{Co-regularization} evaluated their proposed co-regularization method on the $\text{WebKB}_{4}$, and compared it to the single-view regularization method, single-view SVM and co-trained Laplace SVM. \cite{yu2011bayesian} illustrated the co-training algorithm in a graphical way, developed Bayesian co-training, and performed experiments on the $\text{WebKB}_{5}$. On the $\text{WebKB}_{6}$, \cite{dr_cca_pca} compared the performances of multi-view approaches and single-view approaches in respect of subspace learning.

\cite{gonen2008localized, varma2009more, rakotomamonjy2008simplemkl} and \cite{xu2010simple} used the benchmark datasets from the UCI machine learning repository. Thus we use $\text{UCI}_{1},\cdots, \text{UCI}_{4}$ to denote the respective different experiments of these works. In these experiments, several representative multiple kernel learning methods, such as localized MKL and simple MKL, were evaluated in terms of accuracy and time cost. From these comparison results, we discover that multi-view learning methods designed appropriately for real-world applications can indeed improve performance significantly compared to single-view learning methods.
\begin{table*}
\tiny
\centering
\caption{Comparison between multi-view learning and single-view learning methods}\label{tab:result_co-training}
\vspace{0.1in}
\begin{tabular}{|l|c|c|c|c|c|}
  \hline
  DataSet (reference) & Data & \multicolumn{2}{|c|}{Single-view} & \multicolumn{2}{|c|}{Multi-view} \\\hline
  $\text{WebKB}_{1}$  &   & \textbf{naive Bayes} & & \textbf{Co-trained NB} &  \\\cline{2-6}
  \citep{Co-training} & Page & $12.9\%$ & & $6.2\%$ &  \\\cline{2-6}
  Error rate & Hyperlink & 12.4\% & & 11.6\% &  \\\cline{2-6}
   & Page+Hyperlink & 11.1\% & & 5.0\% &  \\\hline
\hline
  $\text{WebKB}_{2}$  &   & \textbf{naive Bayes} & & \textbf{Co-trained NB} & \textbf{Co-EM NB}  \\\cline{2-6}
  \citep{Co-em}  & Page+Hyperlink & 13.0\% & & 5.4\% & 4.3\%\\
  Error rate && & &  &   \\\hline
\hline
 $\text{WebKB}_{3}$  &   & \textbf{naive Bayes}&\textbf{SVM} & \textbf{Co-EM NB} & \textbf{Co-EM SVM}  \\\cline{2-6}
  \citep{co_em_svm}  & Page+Hyperlink & 13.0\% & 10.39\% & 5.08\% & 0.99\% \\
  Error rate && & &  &   \\\hline
  \hline
  $\text{WebKB}_{4}$  &   & \textbf{SVM}&\textbf{RLS} & \textbf{Co-LapSVM} & \textbf{Co-LapRLS}  \\\cline{2-6}
 \citep{Co-regularization}   & Page & 77.8\% & 71.6\% & 93.3\% & 92.0\%  \\\cline{2-6}
 mean PRBEP & Hyperlink & 74.4\% & 72.0\%  & 94.3\% & 94.4\%  \\\cline{2-6}
  & Page+Hyperlink & 84.4\% & 78.3\% & 94.2\% & 93.6\% \\\hline
  \hline
  $\text{WebKB}_{5}$  &   & \textbf{GPLR}& & \textbf{Co-trained GPLR} & \textbf{Bayesian Co-training}  \\\cline{2-6}
  \citep{yu2011bayesian}  & Page+Hyperlink & 0.57\% & & 0.56 \%& 0.58\% \\
 AUC && & &  &   \\\hline
  \hline
  $\text{WebKB}_{6}$  &   & \textbf{KPCA}& & \textbf{KCCA} & \textbf{MKCCA}  \\\cline{2-6}
  \citep{dr_cca_pca}  & Page+Hyperlink & 94.5\% & & 86.6\% & 94.6\% \\
  AUC && & &  &   \\\hline
  \hline
   $\text{UCI}_{1}$  &   & \textbf{SVM} $\text{K}_{P}$ & & \textbf{MKL} $\text{K}_{P}-\text{K}_{G}$ & \textbf{LMKL} $\text{K}_{P}-\text{K}_{G}$  \\\cline{2-6}
  \citep{gonen2008localized}  & Banana & 56.51\% &  & 81.99\% & 83.84\%  \\\cline{2-6}
    ACC & Heart & 72.78\% &  & 75.78\% & 79.44\%  \\\cline{2-6}
  & Ionosphere & 91.54\% &  & 93.68\% & 93.33\%  \\\cline{2-6}
  & Pima & 66.95\%  & & 98.86\% & 98.69\% \\\cline{2-6}
  & Sonar & 65.29\% &  & 80.29\% & 79.57\%\\\hline
\hline
   $\text{UCI}_{2}$  &   & \textbf{LP-SVM} & & \textbf{MKL} & \textbf{GMKL}  \\\cline{2-6}
 \citep{varma2009more}  & Ionosphere & 93.0\% &  & 87.7\% & 94.1\%  \\\cline{2-6}
 ACC   & Parkinsons & 86.2\% &  & 84.7\% & 92.6\%  \\\cline{2-6}
  & Musk & 81.5\% &  & 87.0\% & 93.3\%  \\\cline{2-6}
  & Sonar & 73.7\%  & & 79.5\% & 82.0\% \\\cline{2-6}
  & Wpbc & 76.2\% &  & 69.4\% & 78.3\%\\\hline
\hline
   $\text{UCI}_{3}$  &   &  & & \textbf{SILP} & \textbf{Simple MKL}  \\\cline{2-6}
  \citep{rakotomamonjy2008simplemkl}  & Liver &  &  & 65.9\%\;(47.6) & 65.9\%\;(18.9)  \\\cline{2-6}
 ACC (Time(s)) & Pima &  &  & 76.5\%\;(224) & 76.5\%\;(79.0)  \\\cline{2-6}
  & Ionosphere &  &  & 91.7\%\;(535) & 91.5\%\;(123)  \\\cline{2-6}
  & Wpbc &  & & 76.8\%\;(88.6) & 76.7\%\;(20.6) \\\cline{2-6}
  & Sonar &  &  & 80.5\%\;(2290) & 80.6\%\;(163)\\\hline
  \hline
   $\text{UCI}_{4}$  &   &  & & \textbf{Simple MKL} & \textbf{MKLGL}  \\\cline{2-6}
 \citep{xu2010simple} & Ionosphere &  &  & 91.5\%\;(79.9) & 92.0\%\;(12.0)  \\\cline{2-6}
  ACC (Time(s)) & Breast &  &  & 96.5\%\;(110.5) & 96.6\%\;(14.1)  \\\cline{2-6}
  & Sonar &  &  & 82.0\%\;(57.0) & 82.0\%\;(5.7)  \\\cline{2-6}
  & Pima &  & & 73.4\%\;(94.5) & 73.5\%\;(15.1) \\\hline
\end{tabular}
\end{table*}

\section{Conclusions}
\label{sec:10}

In many scenarios, more than one view can be provided to describe the data. Instead of selecting one view from the corpus or simply concatenating them for learning, we are more interested in algorithms that can learn models from multi-view data by considering the diversity of different views. In this survey paper, we have therefore reviewed several current trends of multi-view learning and classified these algorithms into three different settings: co-training, multiple kernel learning, and subspace learning. Through analyzing these different approaches to the integration of multiple views, we observe that they mainly depend on either the consensus principle or the complementary principle to ensure their success. Furthermore, we also studied the problems with respect to how to construct multiple views and how to evaluate these views. The experimental results show the extensive development of multi-view learning and its promising performance compared to single-view learning.

Although significant work has been carried out in this field, several important research issues need to be addressed in the future. Since the properties of different views largely influence the performance of multi-view learning, it is necessary to place more emphasis on methods to construct, analyze and evaluate the views. For the three groups of multi-view learning algorithms, each have their own advantages, but they are mainly designed and developed separately. Therefore it would be valuable to develop a general framework of multi-view learning which includes the merits of different multi-view learning methods.

We conclude that multi-view learning is effective and promising in practice, but it has not been well-addressed to date. There is still much work to be done to better process multi-view data in a wide variety of applications.




\vskip 0.2in
\bibliography{refs}

\end{document}